\documentclass{bmvc2k}

\usepackage[utf8]{inputenc}
\usepackage{amsmath,amssymb}
\usepackage{booktabs,multirow,array,adjustbox,makecell}
\usepackage{enumitem}
\usepackage[normalem]{ulem}
\usepackage[font=small,skip=7pt]{caption}
\usepackage{placeins}

\title{Continuity-Driven Representation Learning for Industrial Defect Detection}

\addauthor{Minjong Kim}{mj9327@cau.ac.kr}{1}
\addauthor{Hyun Jun Kim}{hyunjun0615@cau.ac.kr}{1}
\addauthor{Jeongrae Kim}{kjk632@cau.ac.kr}{1}
\addauthor{Heeseung Shin}{hs970416@cau.ac.kr}{1}
\addauthor{Changwon Lim}{clim@cau.ac.kr}{1}

\addinstitution{
Chung-Ang University\\
Seoul, Republic of Korea
}

\runninghead{Kim et al.}
{Continuity-Driven Representation Learning for Industrial Defect Detection}

\begin{document}

\maketitle
 
\begin{abstract}
Industrial defect detection differs from natural-image object detection because inspection images are captured under controlled conditions and contain large normal-dominant regions with repetitive structures. Defects therefore appear as localized disruptions of otherwise predictable patterns, while conventional detectors rely mainly on sparse bounding-box supervision, resulting in weakly constrained normal-region representations. We propose a continuity-driven representation regularization framework that exploits normal-dominant regions as dense auxiliary supervision. The framework introduces two detector-agnostic objectives: Multi-Continuity Loss, which combines 1D patch-sequence prediction and 2D masked spatial prediction, and Differencing Loss, which regularizes first-order feature variation and second-order curvature between neighboring patch embeddings. Both objectives are applied with box-derived region weighting to stabilize normal-region representations while preserving defect-related discontinuities.

Experiments on two real-world industrial datasets and the public NEU-DET benchmark, using six detector architectures including YOLO-family models, MambaYOLO, and DETR, demonstrate consistent improvements over native detector baselines. In the full-data setting, the proposed regularizers improve average mAP@0.5:0.95 by up to 3.49 percentage points on Industrial Metal, 5.38 percentage points on MEA, and 5.03 percentage points on NEU-DET. Under limited-data conditions, the gains become more pronounced, with Differencing Loss achieving improvements of up to 21.07 percentage points in mAP@0.5 and 8.23 percentage points in mAP@0.5:0.95 on NEU-DET using only 25\% of the training data. These results suggest that continuity-driven regularization provides an effective prior for improving industrial defect detection, particularly when annotated data are scarce.

\end{abstract}

\section{Introduction}\label{sec:introduction}

Automated visual inspection is a central problem in industrial quality control, where visual defects must be detected reliably under high-throughput manufacturing conditions \citep{nahar2026ai, kuhlechner2025object, cheng2026realworld}. Although modern object detectors such as Faster R-CNN, RetinaNet, YOLO, DETR, and their variants have been widely adopted for defect localization \citep{ren2015faster, redmon2016yolo, lin2017focal, carion2020detr, zhu2021deformable}, industrial defect detection differs substantially from natural-image object detection. In parallel, industrial anomaly detection methods model normal patterns and identify deviations from them, reducing the dependence on dense defect annotations \citep{roth2022patchcore, defard2020padim, wang2021studentteacher, bergmann2019mvtec, chalapathy2019survey, bergmann2020uninformed}. However, these anomaly detection approaches are often not directly optimized for supervised bounding-box localization, whereas object detectors rely mainly on sparse defect-box supervision. This gap motivates our goal: retaining the localization capability of object detectors while exploiting normal-region structure as an additional source of representation-level supervision.

Natural image benchmarks such as MS COCO, PASCAL VOC, and Open Images contain diverse object categories, backgrounds, poses, and semantic contexts \citep{lin2014coco, everingham2010pascal, kuznetsova2020openimages}, whereas industrial inspection images are often acquired under fixed viewpoints, controlled illumination, and predefined regions of interest \citep{steger2018machine,kuhlechner2025object,cheng2026realworld}. As a result, most pixels correspond to normal structures that are repetitive and spatially regular, while defects are typically small, localized, and visually subtle. Figure 1 illustrates this distinction: while natural-image detection focuses on semantic object instances under diverse contexts, industrial defect detection often requires identifying localized structural violations within repetitive normal patterns.

\begin{figure}[t]
\centering
\begin{tabular}{cc}
\includegraphics[width=0.48\textwidth]{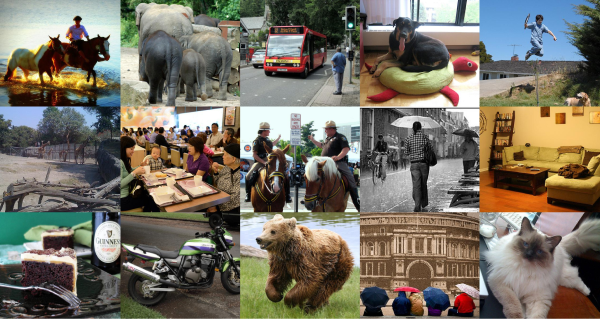} &
\includegraphics[width=0.48\textwidth]{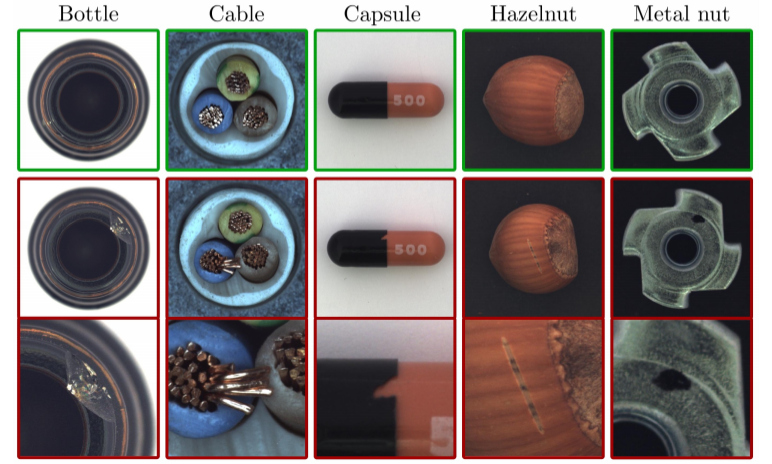} \\
(a) Natural images & (b) Industrial inspection images \\[4pt]
\end{tabular}
\caption{Comparison between natural-image object detection and industrial defect detection. Industrial inspection images contain repetitive normal structures, where defects appear as localized disruptions of structural continuity.}
\label{fig:domain-gap}
\end{figure}
Industrial defects should not be regarded only as independent object instances. In many inspection scenarios, defects appear as localized disruptions of otherwise predictable normal patterns, such as scratches, contamination, or cracks on repetitive structures. Therefore, effective defect detection requires not only learning discriminative defect features, but also learning stable representations of the surrounding normal-dominant regions.
However, conventional object detectors rely mainly on sparse bounding-box supervision, which provides limited constraints on normal-region representations occupying most of the image. Our continuity objectives complement sparse, defect-centric box supervision with dense feature-level constraints over normal-dominant regions, while box-derived weighting prevents defect-related discontinuities from being over-regularized. Our approach specifically targets supervised industrial inspection settings where normal regions exhibit repetitive and relatively homogeneous structures.
To address this issue, we propose a continuity-driven representation regularization framework for industrial defect detection. Our key idea is to exploit normal-dominant regions as dense auxiliary supervision by encouraging predictable patch-level feature transitions while down-weighting annotated defect regions to preserve defect-related discontinuities. The proposed regularizers are applied to intermediate feature maps during training without modifying the detector architecture or native detection objectives.
We introduce two independent auxiliary objectives. Multi-Continuity Loss combines 1D patch-sequence prediction and 2D masked spatial prediction to model ordered representation flow and local spatial consistency. Differencing Loss regularizes first-order feature variation and second-order curvature between neighboring patch embeddings to stabilize structural representation trends in normal regions. Both losses employ box-derived region weighting so that continuity regularization is mainly enforced on normal-dominant areas.

Our contributions are summarized as follows:

\begin{enumerate}
\def\labelenumi{\arabic{enumi}.}
\item
  We formulate supervised industrial defect detection from a structural-continuity perspective, where normal regions form predictable feature fields and defects are treated as localized violations of this continuity.
\item
  We propose Multi-Continuity Loss, a detector-agnostic auxiliary objective that combines 1D patch-sequence prediction and 2D masked spatial prediction for continuity-aware feature regularization.
\item
  We introduce Differencing Loss, which regularizes first-order feature variation and second-order curvature to stabilize normal-region representation trends.
\item
  We incorporate box-derived region weighting and validate the proposed regularizers across two real-world industrial defect datasets, six detector architectures, and data-scarce training settings.
\end{enumerate}

\section{Related Work}\label{related-work}

\subsection{Industrial Defect Detection and Normality Modeling}\label{industrial-defect-detection-and-normality-modeling}

Deep learning-based object detectors have been widely adopted for supervised industrial defect localization because they directly predict bounding boxes and class labels for defective regions \citep{kuhlechner2025object, cheng2026realworld}. CNN-based detectors, including Faster R-CNN, YOLO, and RetinaNet, formulate detection through classification and localization objectives, while Transformer-based detectors such as DETR and Deformable DETR further improve global context modeling through attention mechanisms \citep{ren2015faster, redmon2016yolo, lin2017focal, carion2020detr, zhu2021deformable}. However, industrial inspection images differ substantially from natural-image datasets because they are typically acquired under controlled conditions and contain repetitive normal structures \citep{steger2018machine, kuhlechner2025object, cheng2026realworld}. As a result, defects often appear as localized disruptions of otherwise predictable normal patterns, while detector supervision remains sparse and concentrated mainly on annotated defect regions.

Industrial anomaly detection models normality through distribution- or memory-based representations \citep{defard2020padim, roth2022patchcore} and student--teacher feature matching \citep{wang2021studentteacher, bergmann2020uninformed}, while recent methods have also explored synthetic or diffusion-based anomaly generation \citep{zhang2023diffusion}. These approaches primarily target anomaly scoring or localization maps, whereas our framework retains the native supervised bounding-box objective and uses continuity regularization as auxiliary representation-level supervision.

\subsection{Patch-Level Representation and Spatial Consistency Learning}
\label{patch-level-representation-and-spatial-consistency-learning}

Patch-level representation learning has become an important component of industrial anomaly detection and inspection because industrial defects often appear as localized disruptions of repetitive normal structures \citep{bergmann2019mvtec, chalapathy2019survey, li2025survey}. Representative methods such as PaDiM, PatchCore, and ReConPatch model normal-region representations or feature consistency for anomaly localization \citep{roth2022patchcore, defard2020padim, hyun2024reconpatch}. More recent approaches further introduce spatial-consistency regularization to improve anomaly localization performance \citep{kim2025space}. Smoothness priors such as total variation (TV) and Laplacian regularization have been widely used to suppress local fluctuations in images or feature maps. However, uniform smoothness may suppress localized discontinuities that provide important defect evidence. Our Differencing Loss is related to such smoothness regularization but differs in two aspects. First, it jointly regularizes first- and second-order feature variations to capture local transitions and their curvature. Second, it employs box-derived region weighting to selectively regularize normal-dominant regions while preserving defect-related discontinuities. Detailed comparisons with TV and Laplacian regularization across different training-data ratios are provided in Appendix A. Differencing Loss generally improves detection performance over these conventional smoothness regularizers, particularly under data-scarce settings.

Unlike existing patch-level approaches primarily designed for anomaly scoring or anomaly-map generation, our framework targets supervised bounding-box localization. It retains the native detection objective and applies continuity-driven auxiliary regularization to intermediate feature maps, enabling existing detectors to exploit structural continuity without modifying the detector architecture.

\subsection{Data-Efficient Industrial Inspection}
\label{data-efficient-industrial-inspection}

Industrial defect datasets often contain limited annotated defects because defective samples are rare and costly to collect \citep{bergmann2019mvtec, chalapathy2019survey, bergmann2020uninformed, li2025survey, cheng2026realworld}. Prior approaches have explored data augmentation and synthetic anomaly generation to improve generalization under limited-data settings \citep{shorten2019augmentation, zhang2018mixup, yun2019cutmix}. However, unrealistic transformations or synthetic anomalies may not accurately reflect real industrial defects or imaging conditions \citep{steger2018machine, cheng2026realworld}.

Instead of generating additional defect samples, our framework exploits normal-dominant regions already present in labeled detection images. The proposed continuity regularization provides dense auxiliary supervision from local feature relationships, improving representation stability under limited-data settings without requiring synthetic anomalies or additional labels.

\section{Method}
\label{method}

\subsection{Overall Framework}
\label{overall-framework}

Let $f_{\theta}$ denote a baseline object detector and $\mathcal{L}_{\det}$ its native detection loss. The form of $\mathcal{L}_{\det}$ depends on the detector architecture, such as dense detection losses for YOLO-family detectors and Hungarian matching-based losses for DETR. We do not modify the detector architecture, detection head, or detector-specific objective. Instead, we add an auxiliary feature-level regularizer to intermediate spatial representations during training.

\begin{figure}[t]
\centering
\includegraphics[width=0.8\linewidth]{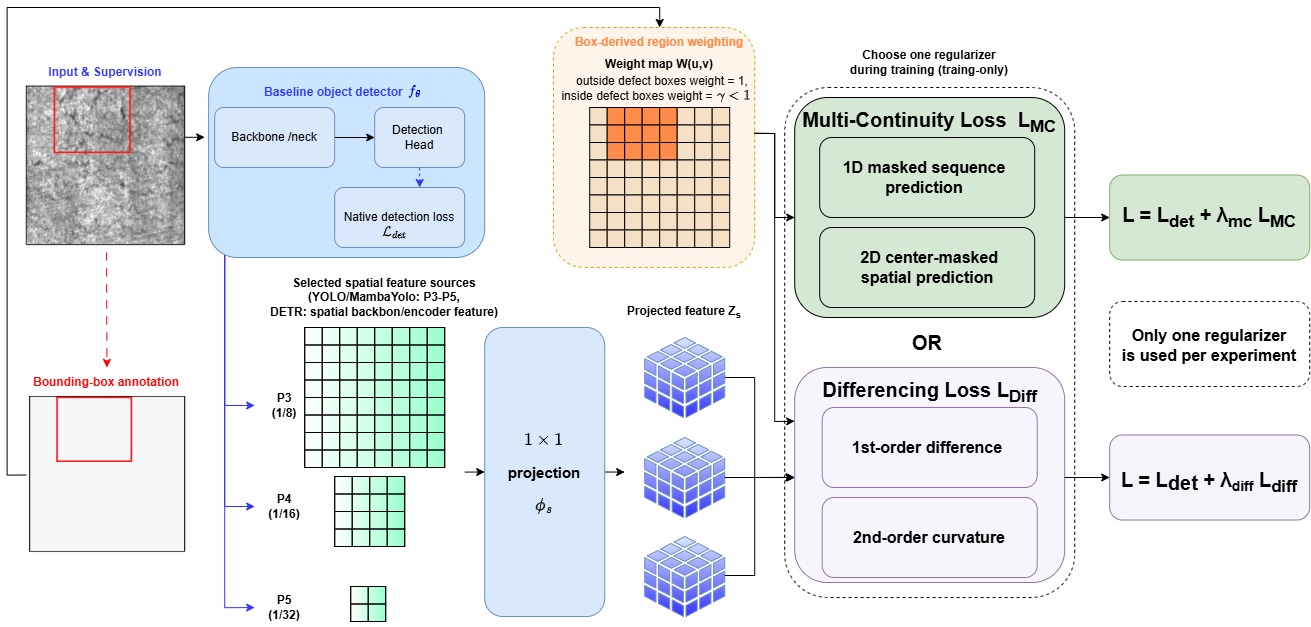}
\caption{Overall framework of the proposed continuity-driven representation regularization. Intermediate detector features are regularized using Multi-Continuity Loss or Differencing Loss with box-derived region weighting during training.}
\label{fig:overall-framework}
\end{figure}

Figure~\ref{fig:overall-framework} illustrates the proposed framework, where intermediate detector features are regularized using either Multi-Continuity Loss or Differencing Loss. The auxiliary modules are used only during training and introduce no inference-time overhead.

Let $\mathcal{S}$ be the set of selected feature sources. For each source $s \in \mathcal{S}$, we denote the corresponding spatial feature map by
\[
F^{s} \in \mathbb{R}^{B \times C_{s} \times H_{s} \times W_{s}},
\]
where $B$, $C_{s}$, $H_{s}$, and $W_{s}$ denote the batch size, channel dimension, height, and width, respectively. A lightweight projection layer $\phi_{s}$, implemented as a $1 \times 1$ convolution, maps each feature map to a common embedding space:
\[
Z^{s} = \phi_{s}(F^{s}), \quad
Z^{s} \in \mathbb{R}^{B \times D \times H_{s} \times W_{s}},
\]
where $D$ is the embedding dimension used for the proposed regularization losses. The projected feature map $Z^{s}$ is used only for auxiliary regularization and is discarded during inference.

We consider two independent regularization variants: Multi-Continuity Loss and Differencing Loss. They are not optimized jointly in our main experiments. For the Multi-Continuity variant, the training objective is
\[
\mathcal{L}
=
\mathcal{L}_{\det}
+
\lambda_{mc}\mathcal{L}_{MC},
\]
whereas for the Differencing variant, the objective is
\[
\mathcal{L}
=
\mathcal{L}_{\det}
+
\lambda_{diff}\mathcal{L}_{Diff}.
\]

Both auxiliary losses are applied with the same box-derived region weighting scheme, described in Section~3.4, to emphasize normal-dominant regions while down-weighting annotated defect boxes.

\subsection{Multi-Continuity Loss}\label{multi-continuity-loss}

The Multi-Continuity Loss models feature continuity from two complementary perspectives. The first term models feature transitions along a one-dimensional patch sequence derived from a two-dimensional feature map, while the second term preserves local spatial adjacency through masked-convolutional prediction in the original feature map domain.

For each selected feature source \(s \in \mathcal{S}\), the projected feature map \(Z^{s}\) is used to compute both 1D and 2D continuity terms.

\subsubsection{1D masked sequence prediction}\label{d-masked-sequence-prediction}

For each projected feature map \(Z^{s}\), we flatten the spatial dimensions into a patch sequence:

\[X_{s} = \left[ x_{1},x_{2},\cdots x_{N} \right],\ \ N = H_{s}W_{s},\]

where \(x_{i} \in \mathbb{R}^{D}\) denotes the embedding of the \(i\)-th spatial location under a fixed raster-scan ordering. This ordering is used only as a computational device for local context prediction; we do not assume that images possess an intrinsic temporal order.

A lightweight one-dimensional predictor \(g_{1D}\) estimates each patch embedding from its masked local sequence context:

\[{\widehat{x}}_{i} = g_{1D}(\mathcal{C}_{i}^{1D}),\]

where \(\mathcal{C}_{i}^{1D}\) denotes the allowed neighboring context around location \(i\). The location-wise discrepancy is defined as

\[d_{i}^{1D} = \delta({\widehat{x}}_{i},x_{i}),\]

where \(\delta( \cdot , \cdot )\) is a feature discrepancy measure such as \(L_{1}\) or cosine distance. This term encourages predictable feature transitions in normal regions.
Collecting the discrepancies over all patch locations, we obtain a location-wise 1D continuity discrepancy sequence

\[\mathbf{d}_{s}^{1D} = \left[ d_{1}^{1D},d_{2}^{1D},\ldots,d_{N}^{1D} \right].\]

For compatibility with the spatial weight map, this sequence can be reshaped back to the feature-map resolution:

\[D_{s}^{1D} \in \mathbb{R}^{H_{s} \times W_{s}}.\]

\subsubsection{2D center-masked spatial prediction}\label{d-center-masked-spatial-prediction}

Although the 1D formulation captures ordered feature transitions, flattening may weaken the original 2D spatial structure. To address this, we introduce a 2D masked-convolutional prediction term.
Given the projected feature map \(Z^{s} \in \mathbb{R}^{B \times D \times H_{s} \times W_{s}}\), we predict each embedding from its local spatial neighborhood using a center-masked convolutional predictor \(g_{2D}\), where the center location is excluded from prediction:

\[{\widehat{z}}_{u,v} = g_{2D}(\mathcal{N}_{u,v}^{2D}\backslash z_{u,v})\]

\[d_{u,v}^{2D} = \delta\left( {\widehat{z}}_{u,v},z_{u,v} \right).\]

The corresponding discrepancy is defined as follows, and all location-wise discrepancies form the 2D continuity discrepancy map \(D_{s}^{2D}\).

Unlike generic pairwise smoothness, this term encourages each normal-region representation to be predictable from its local spatial context without enforcing neighboring features to become identical. The 1D term models ordered feature transitions, while the 2D term preserves spatial adjacency. Together, they provide complementary continuity constraints. The final weighted Multi-Continuity Loss is defined in Section 3.4 using box-derived region weighting.

\subsection{Differencing Loss}\label{differencing-loss}

The Differencing Loss is a separate continuity-driven regularization strategy. Unlike Multi-Continuity Loss, which regularizes feature representations through context-based prediction, Differencing Loss directly constrains the variation structure of neighboring patch embeddings. It is motivated by the observation that normal-dominant regions in industrial inspection images often exhibit stable feature transitions, whereas defects or irregular structures can introduce abrupt changes in local representation trajectories.

For each selected feature source \(s \in \mathcal{S}\), we use the same flattened patch sequence

\[X_{s} = \left[ x_{1},x_{2},\cdots x_{N} \right],\ \ N = H_{s}W_{s},\]

defined in Section 3.2. The raster-scan ordering is used only as a computational device for defining local feature transitions, rather than as an assumption that images have an intrinsic temporal structure.

\subsubsection{First-order feature variation}\label{first-order-feature-variation}

The first-order difference measures the local change between adjacent patch embeddings:

\[\Delta x_{i} = x_{i + 1} - x_{i},\ \ i = 1,\ldots,N - 1.\]

The corresponding location-wise first-order variation term is defined as

\[d_{i}^{(1)} = \left\| \Delta x_{i} \right\|_{1}\]

or, alternatively, by a cosine-based feature discrepancy. This term discourages abrupt feature changes in normal-dominant regions and encourages neighboring patch representations to follow stable local transitions.

\subsubsection{Second-order curvature variation}\label{second-order-curvature-variation}

While the first-order term controls the magnitude of local feature changes, it does not explicitly constrain how these changes evolve across neighboring locations. We therefore introduce a second-order difference:

\[\Delta^{2}x_{i} = x_{i + 1} - 2x_{i} + x_{i - 1},\ \ i = 2,\ldots,N - 1.\]

The corresponding second-order variation term is

\[d_{i}^{(2)} = \left\| \Delta^{2}x_{i} \right\|_{1}\]

This term penalizes unstable curvature in feature trajectories and encourages consistent variation patterns across normal-dominant regions. Unlike uniform smoothing, the proposed Differencing Loss uses box-derived region weighting to stabilize normal regions while down-weighting annotated defect areas, preventing excessive smoothing of defect-related discontinuities. The first-order and second-order terms define location-wise discrepancy sequences \(d_{s}^{(1)}\) and \(d_{s}^{(2)}\), and the final weighted loss is formulated in Section 3.4 using the same weighting scheme as Multi-Continuity Loss.

\subsection{Box-Derived Region Weighting}
\label{box-derived-region-weighting}

The proposed regularizers are intended to stabilize normal-dominant representations. However, defects themselves often appear as local violations of normal continuity, and applying continuity constraints uniformly may suppress defect-sensitive evidence. We therefore construct a box-derived weight map from the available bounding-box annotations.

For each image, spatial locations outside annotated defect boxes are assigned weight $1$, while locations inside defect boxes are assigned a lower weight $\gamma \in [0,1]$:
\[
W(u,v) =
\begin{cases}
\gamma, & (u,v) \in \mathcal{B}, \\
1, & \text{otherwise},
\end{cases}
\]
where $\mathcal{B}$ denotes the union of annotated defect boxes. The weight map is resized to the spatial resolution of each selected feature map. Since no boundary dilation is applied, this scheme should be interpreted as down-weighting annotated defect regions rather than explicitly removing defect boundaries.

For a two-dimensional discrepancy map $D_s$, the weighted aggregation is
\[
\operatorname{Agg}(D_s,W_s)
=
\frac{\sum_{u,v} W_s(u,v)D_s(u,v)}
{\sum_{u,v} W_s(u,v) + \epsilon},
\]
where $\epsilon$ is a small constant for numerical stability. For sequence-based losses, $W_s$ is flattened using the same raster-scan ordering as the patch sequence. For first-order differences, the weight associated with the target or ending location is used; for second-order differences, the weight associated with the center location is used. This aligns the weight sequence with the corresponding discrepancy terms.

Using this aggregation operator, the final Multi-Continuity Loss is defined as
\[
\mathcal{L}_{MC}
=
\sum_{s \in \mathcal{S}}
\left(
\beta_{1D}\operatorname{Agg}(D_s^{1D},W_s)
+
\beta_{2D}\operatorname{Agg}(D_s^{2D},W_s)
\right),
\]
where $D_s^{1D}$ and $D_s^{2D}$ denote the location-wise 1D and 2D continuity discrepancy maps, and $\beta_{1D}$ and $\beta_{2D}$ control their relative contributions.

Similarly, the final Differencing Loss is defined as
\[
\mathcal{L}_{Diff}
=
\sum_{s \in \mathcal{S}}
\left(
\alpha_{1}\operatorname{Agg}(d_s^{(1)},W_s)
+
\alpha_{2}\operatorname{Agg}(d_s^{(2)},W_s)
\right),
\]
where $d_s^{(1)}$ and $d_s^{(2)}$ denote the first- and second-order discrepancy terms, and $\alpha_{1}$ and $\alpha_{2}$ control their contributions.

This region weighting distinguishes our method from global feature smoothing. Normal-dominant regions receive stronger continuity constraints, while annotated defect regions are regularized less aggressively. As a result, the proposed losses stabilize predictable normal feature fields without enforcing uniform smoothness over defect-sensitive regions.

\paragraph{Robustness to box-scale variation.}
To evaluate robustness to annotation imprecision, we vary only the size of the auxiliary mask boxes derived from the ground-truth annotations while keeping all other training settings fixed. As shown in Table~\ref{tab:box-scale-robustness}, Differencing Loss remains effective across a range of box scales, although performance varies non-monotonically with mask size. In particular, enlarged masks maintain strong performance, with the best performance obtained at $1.5\times$. Competitive performance is also maintained at smaller scales, indicating that the proposed regularization is not overly sensitive to moderate variations in the auxiliary mask size. Changing the mask scale mainly controls how much surrounding normal context is down-weighted during regularization.

\subsection{Implementation Across Detector Architectures}\label{implementation-across-detector-architectures}

The proposed regularizers are designed to be detector-agnostic while requiring intermediate representations that preserve spatial adjacency. Therefore, the auxiliary objectives are applied only to spatial feature maps or spatially arranged tokens, excluding unordered representations such as DETR object query embeddings.
For YOLO-family detectors and MambaYOLO, the proposed losses are applied to multi-scale feature maps (P3–P5) used as inputs to the detection head. These feature maps capture defect-related information at different spatial resolutions and provide suitable representations for continuity regularization. Although the internal layer indices differ across detector implementations, the same principle of selecting spatial representations before the detection head is consistently maintained. The original detection heads and native training objectives remain unchanged.
For DETR, the regularizers are not applied to object query embeddings because they do not preserve explicit spatial neighborhood relationships. Instead, the auxiliary losses are applied to backbone or encoder-stage spatial representations that maintain the two-dimensional structure of the input image. This ensures that the continuity assumption is enforced only on representations where local neighborhood relationships remain meaningful.
In all detector architectures, the original detection objectives are preserved, and the proposed losses are used only as auxiliary training objectives. Since the regularizers are removed during inference, the framework introduces no additional inference-time computation or parameters. Multi-Continuity Loss and Differencing Loss are evaluated independently rather than jointly optimized within the same training run.

\section{Experiments}\label{experiments}

\subsection{Experimental Setup}\label{experimental-setup}

We evaluate the proposed continuity-driven representation regularization framework on six detector architectures:
YOLOv8 \citep{ultralytics2023yolov8},
YOLOv10 \citep{wang2024yolov10},
YOLO11 \citep{ultralytics2024yolo11},
YOLOv12 \citep{tian2025yolov12},
MambaYOLO \citep{wang2024mambayolo},
and DETR \citep{carion2020detr}. Experiments compare three training settings: the native detector baseline, baseline + Multi-Continuity Loss, and baseline + Differencing Loss. The two proposed losses are evaluated independently and are not jointly optimized. For all detectors, the original architecture, detection head, and native detection objective are preserved, and the proposed losses are introduced only during training as auxiliary feature-level regularizers.

For YOLO-family detectors and MambaYOLO, the losses are applied to multi-scale detector-head input features, while for DETR they are applied to spatial backbone or encoder representations instead of unordered object queries. Auxiliary projection and prediction modules are removed during inference and therefore introduce no additional inference-time computation.

Hyperparameters are selected using the validation set. Unless otherwise specified, $\gamma = 0.3$ is used for Differencing Loss. Unless otherwise stated, feature discrepancies are computed using the $L_1$ distance throughout all experiments. All results are reported on the held-out test set.

To evaluate robustness under limited annotated data, we additionally conduct reduced-data experiments using 75\%, 50\%, and 25\% of the original training set while keeping validation and test splits fixed. Unless otherwise stated, all detectors use the same preprocessing and evaluation protocol within each dataset.

\subsection{Datasets and Evaluation Metrics}\label{datasets-and-evaluation-metrics}

We evaluate the proposed framework on two real-world industrial defect detection datasets: an Industrial Metal surface inspection dataset and a membrane-electrode assembly (MEA) inspection dataset. Both datasets were collected under controlled industrial inspection conditions and contain large normal-dominant regions with localized defects, making them suitable for evaluating continuity-driven representation regularization.

The Industrial Metal dataset contains 605 images (484 train / 60 val / 61 test) with an input resolution of \(1280 \times 1280\). Defect sizes range from approximately \(29 \times 29\) to \(258 \times 258\) pixels, corresponding to about \(0.05\%\)–\(4.06\%\) of the image area. The median defect size is approximately \(96 \times 96\) pixels (\(\sim0.56\%\) of the image area), and over \(71\%\) of defects occupy less than \(1\%\) of the image.

The MEA dataset contains 242 images (128 train / 57 val / 57 test). Original inspection images reach resolutions up to \(15{,}260 \times 5{,}453\) before preprocessing. After preprocessing, defect sizes range from approximately \(19 \times 19\) to \(58 \times 58\) pixels, corresponding to about \(0.02\%\)–\(0.21\%\) of the image area, with a median size of approximately \(50 \times 50\) pixels (\(\sim0.16\%\)). All annotated defects occupy less than \(1\%\) of the image area, making this an extreme small-defect detection setting. Compared with the Industrial Metal dataset, the MEA dataset exhibits more homogeneous and repetitive normal structures, which align well with the continuity assumption of the proposed framework.

Representative samples are shown in Figure 2. Validation and test sets are fixed across all experiments, including the reduced-data settings in Section 4.4. We report precision, recall, mAP@0.5, and mAP@0.5:0.95 on the test set. Since industrial defects are highly localized, mAP@0.5:0.95 provides a stricter evaluation of localization quality.

\begin{figure}[t]
\centering
\includegraphics[width=\linewidth]{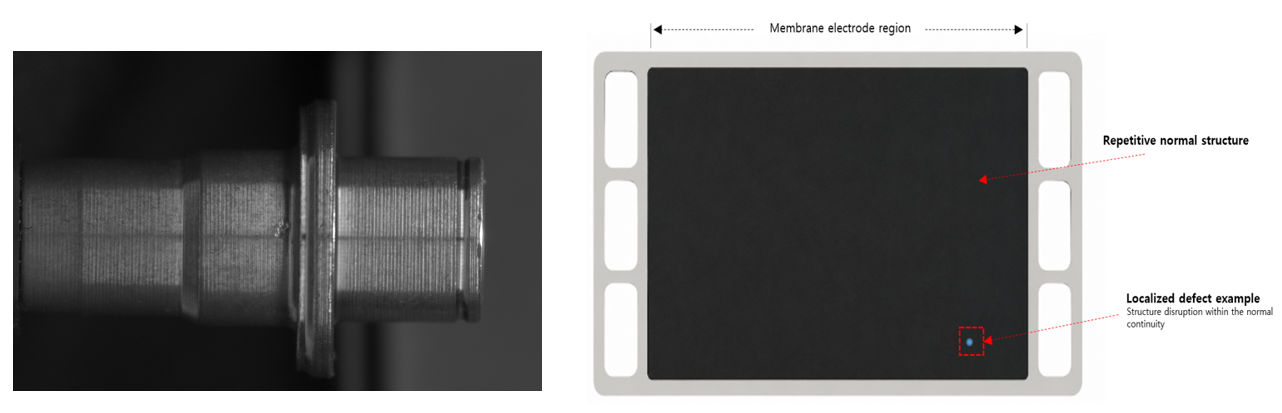}
\caption{Industrial inspection examples used for qualitative illustration. Left: an Industrial Metal inspection image. Right: schematic illustration of the MEA inspection region and localized defect examples. The schematic is used only for visualization and is not used for training or evaluation.
}
\label{fig:dataset-samples}
\end{figure}

\subsection{Main Results}\label{main-results}

\begin{table}[t]
\centering
\caption{Full-data average performance across six detector architectures. Values are averaged over YOLOv8, YOLOv10, YOLO11, YOLOv12, MambaYOLO, and DETR. Improvements are reported in percentage points relative to the corresponding native detector baselines.}
\label{tab:main-full-data-avg}
\footnotesize
\setlength{\tabcolsep}{3pt}
\renewcommand{\arraystretch}{1.08}
\begin{adjustbox}{max width=\textwidth}
\begin{tabular}{llcccc}
\toprule
\textbf{Dataset} & \textbf{Method} & \textbf{mAP@0.5} & \textbf{$\Delta$} & \textbf{mAP@0.5:0.95} & \textbf{$\Delta$} \\
\midrule
\multirow{3}{*}{Industrial Metal}
& Baseline & 0.936 & -- & 0.656 & -- \\
& Multi-Continuity & 0.946 & +1.06 & 0.679 & +2.28 \\
& Differencing & 0.951 & +1.57 & 0.691 & +3.49 \\
\midrule
\multirow{3}{*}{MEA}
& Baseline & 0.993 & -- & 0.890 & -- \\
& Multi-Continuity & 0.994 & +0.15 & 0.942 & +5.19 \\
& Differencing & 0.994 & +0.15 & 0.944 & +5.38 \\
\bottomrule
\end{tabular}
\end{adjustbox}
\end{table}

\begin{table}[t]
\centering
\caption{Variant-level ablation under reduced training data. Values denote average improvements over detector-specific baselines across six detector architectures. All improvements are reported in percentage points. MC denotes Multi-Continuity and Diff. denotes Differencing.}
\label{tab:main-variant-ablation}
\scriptsize
\setlength{\tabcolsep}{3pt}
\renewcommand{\arraystretch}{1.08}
\begin{adjustbox}{max width=\textwidth}
\begin{tabular}{llcccc}
\toprule
\textbf{Dataset} & \textbf{Training ratio} & \makecell{\textbf{MC $\Delta$}\\\textbf{mAP@0.5}} & \makecell{\textbf{Diff. $\Delta$}\\\textbf{mAP@0.5}} & \makecell{\textbf{MC $\Delta$}\\\textbf{mAP@0.5:0.95}} & \makecell{\textbf{Diff. $\Delta$}\\\textbf{mAP@0.5:0.95}} \\
\midrule
\multirow{4}{*}{Industrial Metal}
& 100\% & +1.06 & +1.57 & +2.28 & +3.49 \\
& 75\% & +1.17 & +2.17 & +3.36 & +4.57 \\
& 50\% & +1.86 & +4.70 & +3.78 & +7.14 \\
& 25\% & +3.74 & +7.59 & +3.49 & +6.76 \\
\midrule
\multirow{4}{*}{MEA}
& 100\% & +0.15 & +0.15 & +5.19 & +5.38 \\
& 75\% & +0.33 & +0.97 & +0.35 & +1.51 \\
& 50\% & +3.46 & +4.21 & +4.53 & +2.05 \\
& 25\% & +2.33 & +4.54 & +5.47 & +6.91 \\
\bottomrule
\end{tabular}
\end{adjustbox}
\end{table}

\begin{table}[t]
\centering
\caption{Cross-detector generality analysis. Values denote average improvements over the native detector baselines across both datasets and all training ratios. All improvements are reported in percentage points.}
\label{tab:main-cross-detector}
\footnotesize
\setlength{\tabcolsep}{3pt}
\renewcommand{\arraystretch}{1.08}
\begin{adjustbox}{max width=\textwidth}
\begin{tabular}{lcccc}
\toprule
\textbf{Detector} & \makecell{\textbf{MC $\Delta$}\\\textbf{mAP@0.5}} & \makecell{\textbf{Diff. $\Delta$}\\\textbf{mAP@0.5}} & \makecell{\textbf{MC $\Delta$}\\\textbf{mAP@0.5:0.95}} & \makecell{\textbf{Diff. $\Delta$}\\\textbf{mAP@0.5:0.95}} \\
\midrule
YOLOv8 & +2.02 & +3.10 & +4.50 & +5.45 \\
YOLOv10 & +1.64 & +2.72 & +1.70 & +4.01 \\
YOLO11 & +2.27 & +4.56 & +3.81 & +5.48 \\
YOLOv12 & +1.74 & +3.75 & +4.89 & +5.64 \\
MambaYOLO & +1.10 & +1.62 & +2.41 & +3.35 \\
DETR & +1.81 & +3.68 & +4.04 & +4.41 \\
Average & +1.76 & +3.24 & +3.56 & +4.72 \\
\bottomrule
\end{tabular}
\end{adjustbox}
\end{table}

Table~\ref{tab:main-full-data-avg} summarizes the full-data performance averaged across six detector architectures. Detailed detector-wise results are provided in Appendix A. On the Industrial Metal dataset, both regularization variants improve detection performance over the native detector baselines. Multi-Continuity improves average mAP@0.5 by 1.06 percentage points and mAP@0.5:0.95 by 2.28 percentage points, while Differencing improves mAP@0.5 by 1.57 percentage points and mAP@0.5:0.95 by 3.49 percentage points.

On the MEA dataset, baseline mAP@0.5 is already close to saturation. Nevertheless, both regularizers substantially improve mAP@0.5:0.95. Multi-Continuity improves average mAP@0.5:0.95 by 5.19 percentage points, while Differencing improves it by 5.38 percentage points. This suggests that continuity-driven regularization improves localization quality under stricter IoU thresholds even when coarse detection accuracy is already high.

Overall, the full-data results show that the proposed auxiliary losses improve detector performance without modifying the detector architecture or inference pipeline. Differencing tends to yield larger gains on average, while Multi-Continuity also provides consistent improvements through predictive feature regularization.

\subsection{Variant-Level Ablation under Limited Training Data}\label{variant-level-ablation-under-limited-training-data}

We further analyze the proposed framework under different continuity regularization strategies and training-data regimes. Since Multi-Continuity Loss and Differencing Loss are trained independently, comparisons against the native detector baseline provide a variant-level ablation of the proposed framework.

Table 2 shows that the benefits of continuity regularization become more pronounced as the amount of training data decreases. On the Industrial Metal dataset, Differencing improves average mAP@0.5 from \(+1.57\) percentage points in the full-data setting to \(+7.59\) percentage points at the \(25\%\) training ratio. Similar trends are observed for mAP@0.5:0.95. On the MEA dataset, Multi-Continuity and Differencing improve average mAP@0.5:0.95 by \(+5.47\) and \(+6.91\) percentage points, respectively, under the \(25\%\) training setting. These results indicate that continuity-driven regularization provides effective representation-level supervision when annotated defect samples are limited.

The two regularization variants show complementary behavior. Multi-Continuity provides stable improvements through predictive patch-level continuity modeling, while Differencing often yields larger gains in data-scarce settings by directly constraining local feature variation trends.

Finally, Table 3 summarizes detector-wise average improvements across all datasets and training ratios. Both regularizers consistently improve YOLO-family detectors, MambaYOLO, and DETR, supporting the detector-agnostic nature of the proposed framework.

\subsection{Generalization on Public Industrial Dataset}
\label{generalization-on-public-industrial-dataset}

To further evaluate generalization beyond the two proprietary datasets, we conduct experiments on the public NEU-DET steel surface defect dataset \citep{neu-det}. NEU-DET contains 1,800 images, which we split into training, validation, and test sets using a 6:2:2 ratio. We evaluate the proposed regularizers under two complementary settings. First, we use YOLOv12 at $1280 \times 1280$ to evaluate the method under a high-resolution industrial inspection setting with small localized defects. Second, we conduct a standardized cross-architecture evaluation across six detectors at $640 \times 640$ to assess detector-level generalization. In both settings, improvements are measured relative to the corresponding detector baseline trained under the same resolution and training protocol. We use NEU-DET as an external validation dataset to assess relative improvements rather than to claim benchmark-specific state-of-the-art performance.

We first evaluate YOLOv12 at $1280 \times 1280$ while progressively reducing the amount of training data and keeping the validation and test sets fixed. Table~\ref{tab:main-neudet-ratio-summary} summarizes the improvements over the corresponding YOLOv12 baseline. Multi-Continuity improves performance across all training ratios, with gains of up to 13.97 percentage points in mAP@0.5 and 5.42 points in mAP@0.5:0.95. Differencing provides limited improvement in the full-data setting but becomes substantially more effective as the training data decrease, reaching gains of 21.07 points in mAP@0.5 and 8.23 points in mAP@0.5:0.95 at the 25\% training ratio.

\begin{table}[t]
\centering
\caption{Training-ratio-wise generalization analysis on NEU-DET using YOLOv12 at $1280 \times 1280$ resolution. Values denote improvements over the corresponding YOLOv12 baseline in percentage points.}
\label{tab:main-neudet-ratio-summary}

\footnotesize
\setlength{\tabcolsep}{3pt}
\renewcommand{\arraystretch}{1.08}

\begin{adjustbox}{max width=\textwidth}
\begin{tabular}{lcccc}
\toprule
\makecell{\textbf{Training} \\ \textbf{Ratio}} &
\makecell{\textbf{MC $\Delta$} \\ \textbf{mAP@0.5}} &
\makecell{\textbf{Diff. $\Delta$} \\ \textbf{mAP@0.5}} &
\makecell{\textbf{MC $\Delta$} \\ \textbf{mAP@0.5:0.95}} &
\makecell{\textbf{Diff. $\Delta$} \\ \textbf{mAP@0.5:0.95}} \\
\midrule
100\% & +9.42  & -0.40  & +5.03 & +0.16 \\
75\%  & +7.93  & +16.59 & +3.51 & +7.60 \\
50\%  & +8.43  & +15.16 & +4.30 & +7.27 \\
25\%  & +13.97 & +21.07 & +5.42 & +8.23 \\
\midrule
Average & +9.94 & +13.11 & +4.57 & +5.82 \\
\bottomrule
\end{tabular}
\end{adjustbox}
\end{table}

To examine whether these gains generalize beyond YOLOv12, we further evaluate both regularizers across six detector architectures using a standardized input resolution of $640 \times 640$. Table~\ref{tab:neudet-ratio-avg} reports the average improvement over the corresponding detector-specific baselines. Both variants improve average performance across all training ratios. The gains generally become more pronounced as the amount of training data decreases, particularly for Differencing. At the 25\% training ratio, Multi-Continuity improves mAP@0.5 by 3.75 points, while Differencing achieves gains of 6.23 points in mAP@0.5 and 3.50 points in mAP@0.5:0.95.

\begin{table}[t]
\centering
\caption{Variant-level analysis on NEU-DET across six detector architectures at $640 \times 640$ resolution. Values denote average improvements over detector-specific baselines in percentage points. MC denotes Multi-Continuity and Diff.\ denotes Differencing.}
\label{tab:neudet-ratio-avg}

\scriptsize
\setlength{\tabcolsep}{4pt}
\renewcommand{\arraystretch}{1.08}

\begin{adjustbox}{max width=\linewidth}
\begin{tabular}{lccccc}
\toprule
\textbf{Dataset} &
\textbf{Training ratio} &
\makecell{\textbf{MC $\Delta$} \\ \textbf{mAP@0.5}} &
\makecell{\textbf{Diff. $\Delta$} \\ \textbf{mAP@0.5}} &
\makecell{\textbf{MC $\Delta$} \\ \textbf{mAP@0.5:0.95}} &
\makecell{\textbf{Diff. $\Delta$} \\ \textbf{mAP@0.5:0.95}} \\
\midrule

\multirow{4}{*}{NEU-DET}
& 100\% & +1.75 & +1.60 & +1.63 & +1.73 \\
& 75\%  & +1.15 & +1.12 & +1.68 & +1.27 \\
& 50\%  & +3.25 & +3.37 & +2.35 & +2.05 \\
& 25\%  & +3.75 & +6.23 & +1.78 & +3.50 \\

\bottomrule
\end{tabular}
\end{adjustbox}
\end{table}

Under the full-data setting, the detector baselines already learn relatively discriminative representations, leaving less room for additional regularization. This is particularly evident for YOLOv12 at $1280 \times 1280$, where Differencing slightly decreases mAP@0.5 by 0.40 points while marginally improving mAP@0.5:0.95 by 0.16 points. In contrast, substantially larger gains are observed as the amount of training data decreases. This trend is also observed in the standardized cross-architecture evaluation at $640 \times 640$, where both regularizers yield larger average gains under reduced-data settings. These results indicate that continuity-driven representation regularization generalizes across detector architectures and is particularly effective when annotated training data are limited.

\begin{figure}[!t]
\centering
\includegraphics[width=0.8\linewidth]{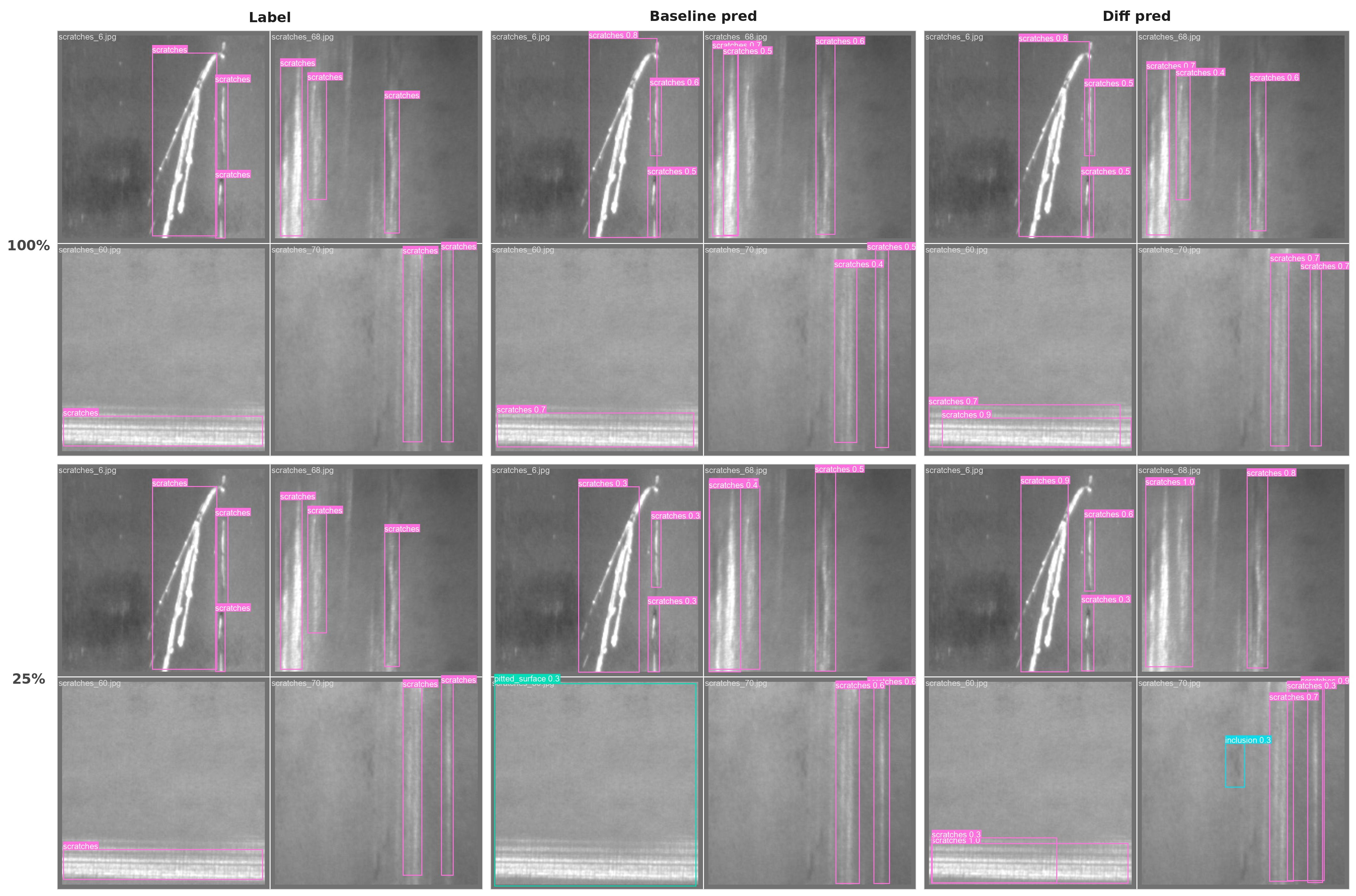}

\caption{Qualitative comparison on NEU-DET under 100\% and 25\% training-data settings. Differencing recovers scratch defects missed by the baseline, while the 25\% setting also shows a detection on an ambiguous defect-like structure.}
\label{fig:neu-det-qualitative}
\end{figure}

Figure~\ref{fig:neu-det-qualitative} further provides qualitative comparisons on NEU-DET under the 100\% and 25\% training-data settings, allowing direct comparison of missed detections, false positives, and localization quality. Differencing recovers scratch defects missed by the baseline under both settings. Under the 25\% setting, an additional detection appears on an ambiguous defect-like structure, indicating that stronger sensitivity under limited supervision may also increase responses to visually similar structures. Together with the quantitative results, these examples illustrate both the benefit and potential limitation of continuity-driven regularization under data-scarce conditions. Detailed detector-wise quantitative results are provided in Appendix~A.

\subsection{Qualitative Analysis}\label{qualitative-analysis}

To further examine how continuity-driven regularization affects learned representations, we visualize intermediate feature maps from YOLOv12 trained with and without the proposed auxiliary objective. Representative feature responses at different depths are provided in Appendix~B, Figure~\ref{fig:feature-representations}. The baseline model is trained using only the standard detection loss, whereas the regularized model incorporates the proposed continuity-based regularization.

\setcounter{figure}{4} 

The qualitative comparison suggests that the proposed regularization does not simply smooth feature representations. Instead, it preserves normal structural patterns while producing more localized and discriminative responses around defect regions. Compared with the baseline, defect-related activations become more distinguishable from the surrounding normal background, indicating improved representational separability between normal and defective regions.

Additional continuity discrepancy visualizations are provided in Appendix~B, Figure~\ref{fig:continuity-discrepancy}. Compared with the baseline, the regularized model exhibits reduced discrepancy fluctuations in homogeneous normal regions and stronger responses around localized structural changes, supporting the effectiveness of continuity-driven representation learning.

This analysis also reveals an important limitation. Strong object boundaries, abrupt geometric transitions, and lighting-induced edges can produce large continuity discrepancies even when they do not correspond to defects. Therefore, continuity discrepancy alone is not sufficient for defect localization. The proposed regularizers should be interpreted as auxiliary representation priors rather than standalone anomaly scores. This observation suggests that the proposed framework is particularly suitable for inspection settings where normal regions exhibit repetitive or homogeneous spatial structures, while additional boundary-aware weighting may be beneficial for objects with complex geometric structures.

\section{Conclusion}\label{conclusion}

We proposed a continuity-driven representation regularization framework for object-detector-based industrial defect detection. Motivated by the observation that industrial defects often appear as localized violations of repetitive normal structures, the proposed framework uses normal-dominant regions as dense auxiliary supervision during detector training. We introduced two detector-agnostic regularizers: Multi-Continuity Loss, which encourages predictable feature transitions through 1D sequence prediction and 2D masked spatial prediction, and Differencing Loss, which constrains first-order feature variation and second-order curvature between neighboring patch embeddings. Both losses are applied with box-derived region weighting to stabilize normal-region representations while avoiding indiscriminate smoothing over annotated defect regions.

Experiments on two real-world industrial inspection datasets, the public NEU-DET benchmark, and six detector architectures show that the proposed regularizers consistently improve supervised defect localization, especially under limited training data. The improvements in mAP@0.5:0.95 indicate that continuity-driven representation learning can enhance localization quality, not merely coarse defect detection. The observed gains on both proprietary industrial datasets and the public NEU-DET dataset further suggest that the proposed regularization framework generalizes across different industrial inspection domains. These results demonstrate that structural continuity serves as an effective representation-level prior for industrial inspection images.

A limitation of the current framework is that normal object boundaries or abrupt geometric transitions may also induce large continuity discrepancies. Future work will investigate boundary-aware weighting, broader public benchmark evaluation, and extensions to anomaly detection and multimodal industrial inspection.


\bibliography{egbib}
\clearpage

\appendix

\setcounter{table}{0}
\renewcommand{\thetable}{A\arabic{table}}
\setcounter{figure}{0}
\renewcommand{\thefigure}{A\arabic{figure}}

\section{Additional Quantitative Results}
\label{app:quantitative-results}

\begin{table}[!t]
\centering
\caption{Robustness to different bounding-box scales on Industrial Metal using YOLOv12 with Differencing Loss.}
\label{tab:box-scale-robustness}

\scriptsize
\setlength{\tabcolsep}{6pt}
\renewcommand{\arraystretch}{1.08}

\begin{tabular}{ccccc}
\toprule
\textbf{Scale} &
\textbf{Precision} &
\textbf{Recall} &
\textbf{mAP@0.5} &
\textbf{mAP@0.5:0.95} \\
\midrule
$0.5\times$ & 0.916 & 0.920 & 0.975 & 0.707 \\
$0.7\times$ & 0.791 & 0.894 & 0.933 & 0.637 \\
$0.9\times$ & 0.905 & 0.825 & 0.936 & 0.618 \\
$1.0\times$ & 0.893 & 0.946 & 0.965 & 0.701 \\
$1.1\times$ & 0.925 & 0.886 & 0.945 & 0.651 \\
$1.3\times$ & 0.934 & 0.875 & 0.971 & 0.723 \\
$1.5\times$ & 0.947 & 0.958 & 0.986 & 0.744 \\
\bottomrule
\end{tabular}
\end{table}

\subsection{Detector-wise Evaluation Results}
\label{app:detector-wise-results}

This appendix reports the detector-wise test results corresponding to the full-data and reduced-data evaluations. The validation and test sets are fixed across all training ratios. For each detector and training ratio, deltas are computed relative to the corresponding detector-specific baseline.

\begin{table}[!htbp]
\centering
\caption{Full-data detector-wise results on Industrial Metal.}
\label{tab:app-industrial-full}
\scriptsize
\setlength{\tabcolsep}{3pt}
\renewcommand{\arraystretch}{1.08}
\begin{adjustbox}{max width=\textwidth}
\begin{tabular}{llcccccc}
\toprule
\textbf{Detector} & \textbf{Method} & \textbf{Precision} & \textbf{Recall} & \textbf{mAP@0.5} & \textbf{mAP@0.5:0.95} & \textbf{$\Delta$mAP@0.5} & \makecell{\textbf{$\Delta$mAP@0.5:0.95}} \\
\midrule
\multirow{3}{*}{YOLOv8}
& Baseline & 0.937 & 0.886 & 0.944 & 0.660 & -- & -- \\
& Multi-Continuity & 0.974 & 0.937 & 0.974 & 0.702 & +2.94 & +4.15 \\
& Differencing & 0.962 & 0.896 & 0.966 & 0.700 & +2.18 & +3.96 \\
\midrule
\multirow{3}{*}{YOLOv10}
& Baseline & 0.875 & 0.861 & 0.922 & 0.661 & -- & -- \\
& Multi-Continuity & 0.765 & 0.920 & 0.927 & 0.663 & +0.51 & +0.15 \\
& Differencing & 0.968 & 0.759 & 0.913 & 0.688 & -0.89 & +2.64 \\
\midrule
\multirow{3}{*}{YOLO11}
& Baseline & 0.985 & 0.848 & 0.945 & 0.659 & -- & -- \\
& Multi-Continuity & 0.891 & 0.890 & 0.937 & 0.669 & -0.76 & +0.97 \\
& Differencing & 0.923 & 0.943 & 0.951 & 0.698 & +0.65 & +3.86 \\
\midrule
\multirow{3}{*}{YOLOv12}
& Baseline & 0.911 & 0.890 & 0.950 & 0.695 & -- & -- \\
& Multi-Continuity & 0.892 & 0.899 & 0.960 & 0.714 & +1.04 & +1.93 \\
& Differencing & 0.893 & 0.946 & 0.965 & 0.701 & +1.47 & +0.63 \\
\midrule
\multirow{3}{*}{MambaYOLO}
& Baseline & 0.937 & 0.890 & 0.953 & 0.659 & -- & -- \\
& Multi-Continuity & 0.945 & 0.860 & 0.956 & 0.683 & +0.32 & +2.37 \\
& Differencing & 0.939 & 0.931 & 0.966 & 0.721 & +1.25 & +6.19 \\
\midrule
\multirow{3}{*}{DETR}
& Baseline & 0.913 & 0.875 & 0.899 & 0.604 & -- & -- \\
& Multi-Continuity & 0.957 & 0.917 & 0.922 & 0.645 & +2.29 & +4.10 \\
& Differencing & 0.944 & 0.944 & 0.947 & 0.640 & +4.77 & +3.65 \\
\bottomrule
\end{tabular}
\end{adjustbox}
\end{table}

\begin{table}[!htbp]
\centering
\caption{Full-data detector-wise results on MEA.}
\label{tab:app-mea-full}
\scriptsize
\setlength{\tabcolsep}{3pt}
\renewcommand{\arraystretch}{1.08}
\begin{adjustbox}{max width=\textwidth}
\begin{tabular}{llcccccc}
\toprule
\textbf{Detector} & \textbf{Method} & \textbf{Precision} & \textbf{Recall} & \textbf{mAP@0.5} & \textbf{mAP@0.5:0.95} & \textbf{$\Delta$mAP@0.5} & \makecell{\textbf{$\Delta$mAP@0.5:0.95}} \\
\midrule
\multirow{3}{*}{YOLOv8}
& Baseline & 0.989 & 1.000 & 0.995 & 0.899 & -- & -- \\
& Multi-Continuity & 0.998 & 0.998 & 0.995 & 0.944 & +0.06 & +4.43 \\
& Differencing & 0.989 & 1.000 & 0.994 & 0.956 & -0.05 & +5.70 \\
\midrule
\multirow{3}{*}{YOLOv10}
& Baseline & 0.989 & 0.998 & 0.995 & 0.901 & -- & -- \\
& Multi-Continuity & 0.997 & 1.000 & 0.995 & 0.960 & +0.00 & +5.90 \\
& Differencing & 0.999 & 0.999 & 0.995 & 0.959 & +0.00 & +5.77 \\
\midrule
\multirow{3}{*}{YOLO11}
& Baseline & 0.982 & 0.998 & 0.994 & 0.898 & -- & -- \\
& Multi-Continuity & 0.999 & 0.998 & 0.994 & 0.952 & +0.05 & +5.39 \\
& Differencing & 0.999 & 0.999 & 0.995 & 0.961 & +0.12 & +6.34 \\
\midrule
\multirow{3}{*}{YOLOv12}
& Baseline & 0.965 & 1.000 & 0.994 & 0.890 & -- & -- \\
& Multi-Continuity & 0.999 & 0.998 & 0.995 & 0.964 & +0.07 & +7.39 \\
& Differencing & 1.000 & 1.000 & 0.995 & 0.943 & +0.07 & +5.34 \\
\midrule
\multirow{3}{*}{MambaYOLO}
& Baseline & 0.966 & 1.000 & 0.990 & 0.877 & -- & -- \\
& Multi-Continuity & 0.998 & 0.997 & 0.994 & 0.902 & +0.44 & +2.53 \\
& Differencing & 0.999 & 0.997 & 0.994 & 0.919 & +0.41 & +4.21 \\
\midrule
\multirow{3}{*}{DETR}
& Baseline & 0.976 & 0.997 & 0.990 & 0.876 & -- & -- \\
& Multi-Continuity & 0.998 & 0.997 & 0.993 & 0.931 & +0.29 & +5.52 \\
& Differencing & 0.992 & 0.998 & 0.994 & 0.925 & +0.37 & +4.90 \\
\bottomrule
\end{tabular}
\end{adjustbox}
\end{table}

\begin{table}[!htbp]
\centering
\caption{Detector-wise results on the Industrial Metal dataset at the 75\% training ratio.}
\label{tab:app-industrial-75}
\scriptsize
\setlength{\tabcolsep}{3pt}
\renewcommand{\arraystretch}{1.08}
\begin{adjustbox}{max width=\textwidth}
\begin{tabular}{llcccccc}
\toprule
\textbf{Method} & \textbf{Detector} & \textbf{Precision} & \textbf{Recall} & \textbf{mAP@0.5} & \textbf{mAP@0.5:0.95} & \textbf{$\Delta$mAP@0.5} & \makecell{\textbf{$\Delta$mAP@0.5:0.95}} \\
\midrule
\multirow{6}{*}{Baseline}
& YOLOv8 & 0.873 & 0.930 & 0.959 & 0.666 & -- & -- \\
& YOLOv10 & 0.859 & 0.951 & 0.965 & 0.645 & -- & -- \\
& YOLO11 & 0.939 & 0.860 & 0.934 & 0.623 & -- & -- \\
& YOLOv12 & 0.900 & 0.861 & 0.931 & 0.612 & -- & -- \\
& MambaYOLO & 0.864 & 0.861 & 0.912 & 0.621 & -- & -- \\
& DETR & 0.924 & 0.847 & 0.872 & 0.564 & -- & -- \\
\midrule
\multirow{6}{*}{Multi-Continuity}
& YOLOv8 & 0.929 & 0.929 & 0.971 & 0.718 & +1.19 & +5.23 \\
& YOLOv10 & 0.887 & 0.919 & 0.951 & 0.642 & -1.45 & -0.26 \\
& YOLO11 & 0.927 & 0.896 & 0.932 & 0.675 & -0.16 & +5.26 \\
& YOLOv12 & 0.912 & 0.901 & 0.936 & 0.657 & +0.52 & +4.48 \\
& MambaYOLO & 0.963 & 0.838 & 0.946 & 0.638 & +3.40 & +1.64 \\
& DETR & 0.969 & 0.875 & 0.908 & 0.603 & +3.53 & +3.84 \\
\midrule
\multirow{6}{*}{Differencing}
& YOLOv8 & 0.905 & 0.915 & 0.953 & 0.698 & -0.66 & +3.21 \\
& YOLOv10 & 0.874 & 0.913 & 0.942 & 0.691 & -2.40 & +4.67 \\
& YOLO11 & 0.942 & 0.937 & 0.977 & 0.681 & +4.28 & +5.84 \\
& YOLOv12 & 0.890 & 0.931 & 0.950 & 0.675 & +1.91 & +6.28 \\
& MambaYOLO & 0.880 & 0.879 & 0.929 & 0.627 & +1.73 & +0.59 \\
& DETR & 0.919 & 0.944 & 0.954 & 0.632 & +8.17 & +6.82 \\
\bottomrule
\end{tabular}
\end{adjustbox}
\end{table}

\begin{table}[!htbp]
\centering
\caption{Detector-wise results on the Industrial Metal dataset at the 50\% training ratio.}
\label{tab:app-industrial-50}
\scriptsize
\setlength{\tabcolsep}{3pt}
\renewcommand{\arraystretch}{1.08}
\begin{adjustbox}{max width=\textwidth}
\begin{tabular}{llcccccc}
\toprule
\textbf{Method} & \textbf{Detector} & \textbf{Precision} & \textbf{Recall} & \textbf{mAP@0.5} & \textbf{mAP@0.5:0.95} & \textbf{$\Delta$mAP@0.5} & \makecell{\textbf{$\Delta$mAP@0.5:0.95}} \\
\midrule
\multirow{6}{*}{Baseline}
& YOLOv8 & 0.792 & 0.872 & 0.881 & 0.527 & -- & -- \\
& YOLOv10 & 0.812 & 0.730 & 0.858 & 0.562 & -- & -- \\
& YOLO11 & 0.787 & 0.883 & 0.904 & 0.603 & -- & -- \\
& YOLOv12 & 0.854 & 0.785 & 0.868 & 0.519 & -- & -- \\
& MambaYOLO & 0.869 & 0.880 & 0.908 & 0.575 & -- & -- \\
& DETR & 0.865 & 0.889 & 0.876 & 0.538 & -- & -- \\
\midrule
\multirow{6}{*}{Multi-Continuity}
& YOLOv8 & 0.874 & 0.914 & 0.933 & 0.635 & +5.18 & +10.75 \\
& YOLOv10 & 0.831 & 0.792 & 0.891 & 0.560 & +3.23 & -0.18 \\
& YOLO11 & 0.825 & 0.838 & 0.891 & 0.567 & -1.34 & -3.62 \\
& YOLOv12 & 0.828 & 0.831 & 0.888 & 0.594 & +2.02 & +7.44 \\
& MambaYOLO & 0.888 & 0.835 & 0.899 & 0.587 & -0.89 & +1.22 \\
& DETR & 0.857 & 0.917 & 0.906 & 0.609 & +2.95 & +7.09 \\
\midrule
\multirow{6}{*}{Differencing}
& YOLOv8 & 0.960 & 0.948 & 0.975 & 0.690 & +9.35 & +16.32 \\
& YOLOv10 & 0.904 & 0.869 & 0.954 & 0.658 & +9.60 & +9.56 \\
& YOLO11 & 0.876 & 0.905 & 0.939 & 0.593 & +3.47 & -1.02 \\
& YOLOv12 & 0.877 & 0.850 & 0.904 & 0.642 & +3.62 & +12.24 \\
& MambaYOLO & 0.907 & 0.809 & 0.900 & 0.595 & -0.75 & +2.08 \\
& DETR & 0.940 & 0.875 & 0.905 & 0.574 & +2.93 & +3.63 \\
\bottomrule
\end{tabular}
\end{adjustbox}
\end{table}

\begin{table}[!htbp]
\centering
\caption{Detector-wise results on the Industrial Metal dataset at the 25\% training ratio.}
\label{tab:app-industrial-25}
\scriptsize
\setlength{\tabcolsep}{3pt}
\renewcommand{\arraystretch}{1.08}
\begin{adjustbox}{max width=\textwidth}
\begin{tabular}{llcccccc}
\toprule
\textbf{Method} & \textbf{Detector} & \textbf{Precision} & \textbf{Recall} & \textbf{mAP@0.5} & \textbf{mAP@0.5:0.95} & \textbf{$\Delta$mAP@0.5} & \makecell{\textbf{$\Delta$mAP@0.5:0.95}} \\
\midrule
\multirow{6}{*}{Baseline}
& YOLOv8 & 0.859 & 0.868 & 0.893 & 0.553 & -- & -- \\
& YOLOv10 & 0.825 & 0.819 & 0.847 & 0.561 & -- & -- \\
& YOLO11 & 0.700 & 0.643 & 0.732 & 0.417 & -- & -- \\
& YOLOv12 & 0.576 & 0.646 & 0.646 & 0.326 & -- & -- \\
& MambaYOLO & 0.812 & 0.716 & 0.797 & 0.405 & -- & -- \\
& DETR & 0.937 & -- & 0.838 & 0.529 & -- & -- \\
\midrule
\multirow{6}{*}{Multi-Continuity}
& YOLOv8 & 0.818 & 0.915 & 0.910 & 0.575 & +1.73 & +2.20 \\
& YOLOv10 & 0.858 & 0.787 & 0.878 & 0.566 & +3.13 & +0.46 \\
& YOLO11 & 0.803 & 0.858 & 0.850 & 0.504 & +11.79 & +8.63 \\
& YOLOv12 & 0.679 & 0.652 & 0.707 & 0.391 & +6.13 & +6.48 \\
& MambaYOLO & 0.744 & 0.766 & 0.784 & 0.421 & -1.31 & +1.54 \\
& DETR & 0.853 & 0.889 & 0.848 & 0.545 & +0.98 & +1.60 \\
\midrule
\multirow{6}{*}{Differencing}
& YOLOv8 & 0.866 & 0.897 & 0.928 & 0.593 & +3.46 & +3.96 \\
& YOLOv10 & 0.838 & 0.815 & 0.878 & 0.546 & +3.16 & -1.54 \\
& YOLO11 & 0.946 & 0.840 & 0.905 & 0.588 & +17.24 & +17.07 \\
& YOLOv12 & 0.698 & 0.847 & 0.790 & 0.461 & +14.38 & +13.55 \\
& MambaYOLO & 0.834 & 0.739 & 0.808 & 0.441 & +1.13 & +3.60 \\
& DETR & 0.984 & 0.847 & 0.900 & 0.568 & +6.18 & +3.92 \\
\bottomrule
\end{tabular}
\end{adjustbox}
\end{table}

\begin{table}[!htbp]
\centering
\caption{Detector-wise results on the MEA dataset at the 75\% training ratio.}
\label{tab:app-mea-75}
\scriptsize
\setlength{\tabcolsep}{3pt}
\renewcommand{\arraystretch}{1.08}
\begin{adjustbox}{max width=\textwidth}
\begin{tabular}{llcccccc}
\toprule
\textbf{Method} & \textbf{Detector} & \textbf{Precision} & \textbf{Recall} & \textbf{mAP@0.5} & \textbf{mAP@0.5:0.95} & \textbf{$\Delta$mAP@0.5} & \makecell{\textbf{$\Delta$mAP@0.5:0.95}} \\
\midrule
\multirow{6}{*}{Baseline}
& YOLOv8 & 0.982 & 0.987 & 0.990 & 0.845 & -- & -- \\
& YOLOv10 & 0.981 & 0.986 & 0.989 & 0.842 & -- & -- \\
& YOLO11 & 0.983 & 0.988 & 0.990 & 0.849 & -- & -- \\
& YOLOv12 & 0.981 & 0.998 & 0.993 & 0.860 & -- & -- \\
& MambaYOLO & 0.945 & 0.952 & 0.951 & 0.810 & -- & -- \\
& DETR & 0.962 & 0.968 & 0.969 & 0.825 & -- & -- \\
\midrule
\multirow{6}{*}{Multi-Continuity}
& YOLOv8 & 0.986 & 0.991 & 0.993 & 0.853 & +0.31 & +0.80 \\
& YOLOv10 & 0.985 & 0.990 & 0.992 & 0.851 & +0.32 & +0.88 \\
& YOLO11 & 0.988 & 0.992 & 0.993 & 0.826 & +0.30 & -2.25 \\
& YOLOv12 & 0.984 & 0.999 & 0.995 & 0.866 & +0.11 & +0.56 \\
& MambaYOLO & 0.952 & 0.958 & 0.957 & 0.821 & +0.52 & +1.13 \\
& DETR & 0.969 & 0.872 & 0.974 & 0.835 & +0.41 & +1.00 \\
\midrule
\multirow{6}{*}{Differencing}
& YOLOv8 & 0.990 & 0.993 & 0.995 & 0.860 & +0.51 & +1.50 \\
& YOLOv10 & 0.988 & 0.992 & 0.995 & 0.859 & +0.59 & +1.66 \\
& YOLO11 & 0.991 & 0.994 & 0.996 & 0.865 & +0.59 & +1.61 \\
& YOLOv12 & 0.981 & 1.000 & 0.994 & 0.864 & +0.04 & +0.35 \\
& MambaYOLO & 0.957 & 0.963 & 0.973 & 0.832 & +2.21 & +2.22 \\
& DETR & 0.972 & 0.975 & 0.988 & 0.842 & +1.88 & +1.71 \\
\bottomrule
\end{tabular}
\end{adjustbox}
\end{table}

\begin{table}[!htbp]
\centering
\caption{Detector-wise results on the MEA dataset at the 50\% training ratio.}
\label{tab:app-mea-50}
\scriptsize
\setlength{\tabcolsep}{3pt}
\renewcommand{\arraystretch}{1.08}
\begin{adjustbox}{max width=\textwidth}
\begin{tabular}{llcccccc}
\toprule
\textbf{Method} & \textbf{Detector} & \textbf{Precision} & \textbf{Recall} & \textbf{mAP@0.5} & \textbf{mAP@0.5:0.95} & \textbf{$\Delta$mAP@0.5} & \makecell{\textbf{$\Delta$mAP@0.5:0.95}} \\
\midrule
\multirow{6}{*}{Baseline}
& YOLOv8 & 0.974 & 0.979 & 0.947 & 0.845 & -- & -- \\
& YOLOv10 & 0.973 & 0.978 & 0.955 & 0.843 & -- & -- \\
& YOLO11 & 0.976 & 0.980 & 0.957 & 0.848 & -- & -- \\
& YOLOv12 & 0.979 & 0.982 & 0.920 & 0.868 & -- & -- \\
& MambaYOLO & 0.941 & 0.947 & 0.894 & 0.812 & -- & -- \\
& DETR & 0.959 & 0.962 & 0.911 & 0.826 & -- & -- \\
\midrule
\multirow{6}{*}{Multi-Continuity}
& YOLOv8 & 0.978 & 0.983 & 0.969 & 0.885 & +2.25 & +4.00 \\
& YOLOv10 & 0.982 & 0.982 & 0.989 & 0.871 & +3.38 & +2.83 \\
& YOLO11 & 0.984 & 0.984 & 0.977 & 0.900 & +2.01 & +5.19 \\
& YOLOv12 & 0.982 & 0.984 & 0.963 & 0.905 & +4.34 & +3.74 \\
& MambaYOLO & 0.966 & 0.952 & 0.953 & 0.868 & +5.81 & +5.59 \\
& DETR & 0.962 & 0.996 & 0.941 & 0.884 & +2.99 & +5.85 \\
\midrule
\multirow{6}{*}{Differencing}
& YOLOv8 & 0.978 & 0.983 & 0.989 & 0.862 & +4.25 & +1.67 \\
& YOLOv10 & 0.978 & 0.982 & 0.989 & 0.859 & +3.38 & +1.62 \\
& YOLO11 & 0.980 & 0.985 & 0.990 & 0.899 & +3.29 & +5.14 \\
& YOLOv12 & 0.980 & 0.982 & 0.991 & 0.860 & +7.11 & -0.75 \\
& MambaYOLO & 0.947 & 0.952 & 0.945 & 0.831 & +5.04 & +1.88 \\
& DETR & 0.966 & 0.966 & 0.933 & 0.853 & +2.17 & +2.70 \\
\bottomrule
\end{tabular}
\end{adjustbox}
\end{table}

\begin{table}[!htbp]
\centering
\caption{Detector-wise results on the MEA dataset at the 25\% training ratio.}
\label{tab:app-mea-25}
\scriptsize
\setlength{\tabcolsep}{3pt}
\renewcommand{\arraystretch}{1.08}
\begin{adjustbox}{max width=\textwidth}
\begin{tabular}{llcccccc}
\toprule
\textbf{Method} & \textbf{Detector} & \textbf{Precision} & \textbf{Recall} & \textbf{mAP@0.5} & \textbf{mAP@0.5:0.95} & \textbf{$\Delta$mAP@0.5} & \makecell{\textbf{$\Delta$mAP@0.5:0.95}} \\
\midrule
\multirow{6}{*}{Baseline}
& YOLOv8 & 0.893 & 0.890 & 0.915 & 0.701 & -- & -- \\
& YOLOv10 & 0.891 & 0.889 & 0.883 & 0.701 & -- & -- \\
& YOLO11 & 0.919 & 0.901 & 0.869 & 0.690 & -- & -- \\
& YOLOv12 & 0.922 & 0.915 & 0.922 & 0.737 & -- & -- \\
& MambaYOLO & 0.885 & 0.899 & 0.904 & 0.699 & -- & -- \\
& DETR & 0.900 & 0.891 & 0.900 & 0.700 & -- & -- \\
\midrule
\multirow{6}{*}{Multi-Continuity}
& YOLOv8 & 0.929 & 0.935 & 0.940 & 0.746 & +2.50 & +4.45 \\
& YOLOv10 & 0.927 & 0.933 & 0.923 & 0.739 & +3.99 & +3.79 \\
& YOLO11 & 0.909 & 0.938 & 0.932 & 0.799 & +6.27 & +10.91 \\
& YOLOv12 & 0.942 & 0.948 & 0.918 & 0.808 & -0.35 & +7.09 \\
& MambaYOLO & 0.905 & 0.925 & 0.910 & 0.731 & +0.55 & +3.24 \\
& DETR & 0.918 & 0.939 & 0.910 & 0.733 & +1.04 & +3.31 \\
\midrule
\multirow{6}{*}{Differencing}
& YOLOv8 & 0.939 & 0.951 & 0.973 & 0.774 & +5.80 & +7.28 \\
& YOLOv10 & 0.943 & 0.949 & 0.966 & 0.778 & +8.31 & +7.70 \\
& YOLO11 & 0.958 & 0.962 & 0.937 & 0.740 & +6.80 & +4.98 \\
& YOLOv12 & 0.965 & 0.959 & 0.936 & 0.812 & +1.42 & +7.49 \\
& MambaYOLO & 0.926 & 0.929 & 0.924 & 0.759 & +1.93 & +6.06 \\
& DETR & 0.935 & 0.931 & 0.930 & 0.779 & +3.00 & +7.92 \\
\bottomrule
\end{tabular}
\end{adjustbox}
\end{table}

\begin{table}[t]
\centering
\caption{Training-ratio-wise results on the NEU-DET dataset using YOLOv12 across 100\%, 75\%, 50\%, and 25\% training ratios.}
\label{tab:app-neudet-YOLOv12-ratio}
\scriptsize
\setlength{\tabcolsep}{3pt}
\renewcommand{\arraystretch}{1.08}
\begin{adjustbox}{max width=\textwidth}
\begin{tabular}{llcccccc}
\toprule
\textbf{Method} & \makecell{\textbf{Training} \ \textbf{Ratio}} & \textbf{Precision} & \textbf{Recall} & \textbf{mAP@0.5} & \textbf{mAP@0.5:0.95} & \textbf{$\Delta$mAP@0.5} & \makecell{\textbf{$\Delta$mAP@0.5:0.95}} \\
\midrule
\multirow{4}{*}{Baseline}
& 100\% & 0.658 & 0.292 & 0.364 & 0.140 & -- & -- \\
& 75\%  & 0.444 & 0.416 & 0.347 & 0.138 & -- & -- \\
& 50\%  & 0.453 & 0.383 & 0.318 & 0.112 & -- & -- \\
& 25\%  & 0.492 & 0.299 & 0.210 & 0.067 & -- & -- \\
\midrule
\multirow{4}{*}{Multi-Continuity}
& 100\% & 0.581 & 0.518 & 0.459 & 0.190 & +9.42 & +5.03 \\
& 75\%  & 0.527 & 0.524 & 0.426 & 0.173 & +7.93 & +3.51 \\
& 50\%  & 0.510 & 0.446 & 0.402 & 0.155 & +8.43 & +4.30 \\
& 25\%  & 0.331 & 0.498 & 0.350 & 0.121 & +13.97 & +5.42 \\
\midrule
\multirow{4}{*}{Differencing}
& 100\% & 0.436 & 0.508 & 0.360 & 0.141 & -0.40 & +0.16 \\
& 75\%  & 0.590 & 0.573 & 0.512 & 0.214 & +16.59 & +7.60 \\
& 50\%  & 0.567 & 0.517 & 0.470 & 0.185 & +15.16 & +7.27 \\
& 25\%  & 0.522 & 0.481 & 0.421 & 0.149 & +21.07 & +8.23 \\
\bottomrule
\end{tabular}
\end{adjustbox}
\end{table}
\begin{table}[t]

\centering

\caption{Comparison with conventional smoothness regularization on the Metal and Secondary Battery datasets using YOLOv12 across different training ratios.}

\label{tab:smoothness-comparison}

\scriptsize

\setlength{\tabcolsep}{4pt}

\renewcommand{\arraystretch}{1.08}

\begin{adjustbox}{max width=\textwidth}

\begin{tabular}{lllcc}

\toprule

\textbf{Dataset} &
\textbf{Method} &
\makecell{\textbf{Training} \\ \textbf{Ratio}} &
\textbf{mAP@0.5} &
\textbf{mAP@0.5:0.95} \\

\midrule

\multirow{12}{*}{Metal}

& \multirow{4}{*}{TV Loss}
& 100\% & 0.9519 & 0.7001 \\
& & 75\%  & 0.9259 & 0.6354 \\
& & 50\%  & 0.8229 & 0.5516 \\
& & 25\%  & 0.7535 & 0.4288 \\

\cmidrule(lr){2-5}

& \multirow{4}{*}{Laplacian Loss}
& 100\% & 0.9528 & 0.7074 \\
& & 75\%  & 0.9204 & 0.6200 \\
& & 50\%  & 0.8480 & 0.5657 \\
& & 25\%  & 0.7701 & 0.4492 \\

\cmidrule(lr){2-5}

& \multirow{4}{*}{Differencing Loss}
& 100\% & 0.9646 & 0.7011 \\
& & 75\%  & 0.9497 & 0.6752 \\
& & 50\%  & 0.9037 & 0.6418 \\
& & 25\%  & 0.7896 & 0.4612 \\

\midrule

\multirow{12}{*}{MEA}

& \multirow{4}{*}{TV Loss}
& 100\% & 0.9938 & 0.9187 \\
& & 75\%  & 0.9910 & 0.8508 \\
& & 50\%  & 0.9562 & 0.7358 \\
& & 25\%  & 0.7800 & 0.5560 \\

\cmidrule(lr){2-5}

& \multirow{4}{*}{Laplacian Loss}
& 100\% & 0.9903 & 0.9151 \\
& & 75\%  & 0.9903 & 0.8599 \\
& & 50\%  & 0.9809 & 0.8384 \\
& & 25\%  & 0.9207 & 0.7778 \\

\cmidrule(lr){2-5}

& \multirow{4}{*}{Differencing Loss}
& 100\% & 0.9950 & 0.9433 \\
& & 75\%  & 0.9938 & 0.8636 \\
& & 50\%  & 0.9912 & 0.8600 \\
& & 25\%  & 0.9358 & 0.8123 \\

\bottomrule

\end{tabular}

\end{adjustbox}

\end{table}

\begin{table}[t]
\centering
\caption{Full-data detector-wise results on NEU-DET.}
\label{tab:app-neudet-100}

\scriptsize
\setlength{\tabcolsep}{3pt}
\renewcommand{\arraystretch}{1.08}

\begin{adjustbox}{max width=\textwidth}
\begin{tabular}{llcccccc}
\toprule
\textbf{Detector} & \textbf{Method} &
\textbf{Precision} & \textbf{Recall} &
\textbf{mAP@0.5} & \textbf{mAP@0.5:0.95} &
\textbf{$\Delta$mAP@0.5} & \textbf{$\Delta$mAP@0.5:0.95} \\
\midrule

\multirow{3}{*}{YOLOv8}
& Baseline & 0.686 & 0.683 & 0.728 & 0.408 & -- & -- \\
& Multi-Continuity & 0.680 & 0.674 & 0.746 & 0.414 & +1.80 & +0.60 \\
& Differencing & 0.679 & 0.677 & 0.742 & 0.430 & +1.40 & +2.20 \\
\midrule

\multirow{3}{*}{YOLOv10}
& Baseline & 0.674 & 0.657 & 0.702 & 0.404 & -- & -- \\
& Multi-Continuity & 0.684 & 0.640 & 0.694 & 0.411 & -0.80 & +0.70 \\
& Differencing & 0.661 & 0.666 & 0.701 & 0.415 & -0.10 & +1.10 \\
\midrule

\multirow{3}{*}{YOLO11}
& Baseline & 0.650 & 0.656 & 0.711 & 0.393 & -- & -- \\
& Multi-Continuity & 0.633 & 0.667 & 0.710 & 0.412 & -0.10 & +1.90 \\
& Differencing & 0.674 & 0.655 & 0.717 & 0.409 & +0.60 & +1.60 \\
\midrule

\multirow{3}{*}{YOLOv12}
& Baseline & 0.661 & 0.656 & 0.708 & 0.393 & -- & -- \\
& Multi-Continuity & 0.638 & 0.683 & 0.702 & 0.398 & -0.60 & +0.50 \\
& Differencing & 0.652 & 0.680 & 0.723 & 0.413 & +1.50 & +2.00 \\
\midrule

\multirow{3}{*}{MambaYOLO}
& Baseline & 0.672 & 0.710 & 0.744 & 0.426 & -- & -- \\
& Multi-Continuity & 0.684 & 0.720 & 0.757 & 0.430 & +1.30 & +0.40 \\
& Differencing & 0.740 & 0.679 & 0.747 & 0.430 & +0.30 & +0.40 \\
\midrule

\multirow{3}{*}{DETR}
& Baseline & 0.527 & 0.506 & 0.509 & 0.257 & -- & -- \\
& Multi-Continuity & 0.667 & 0.531 & 0.598 & 0.314 & +8.90 & +5.70 \\
& Differencing & 0.587 & 0.555 & 0.568 & 0.288 & +5.90 & +3.10 \\
\bottomrule
\end{tabular}
\end{adjustbox}
\end{table}

\begin{table}[t]
\centering
\caption{Detector-wise results on NEU-DET using 75\% of the training data.}
\label{tab:app-neudet-75}

\scriptsize
\setlength{\tabcolsep}{3pt}
\renewcommand{\arraystretch}{1.08}

\begin{adjustbox}{max width=\textwidth}
\begin{tabular}{llcccccc}
\toprule
\textbf{Detector} & \textbf{Method} &
\textbf{Precision} & \textbf{Recall} &
\textbf{mAP@0.5} & \textbf{mAP@0.5:0.95} &
\textbf{$\Delta$mAP@0.5} & \textbf{$\Delta$mAP@0.5:0.95} \\
\midrule

\multirow{3}{*}{YOLOv8}
& Baseline & 0.668 & 0.689 & 0.702 & 0.378 & -- & -- \\
& Multi-Continuity & 0.661 & 0.685 & 0.722 & 0.414 & +2.00 & +3.60 \\
& Differencing & 0.667 & 0.684 & 0.724 & 0.408 & +2.20 & +3.00 \\
\midrule

\multirow{3}{*}{YOLOv10}
& Baseline & 0.638 & 0.630 & 0.676 & 0.372 & -- & -- \\
& Multi-Continuity & 0.669 & 0.638 & 0.688 & 0.395 & +1.20 & +2.30 \\
& Differencing & 0.613 & 0.632 & 0.659 & 0.380 & -1.70 & +0.80 \\
\midrule

\multirow{3}{*}{YOLO11}
& Baseline & 0.605 & 0.678 & 0.693 & 0.386 & -- & -- \\
& Multi-Continuity & 0.664 & 0.626 & 0.700 & 0.387 & +0.70 & +0.10 \\
& Differencing & 0.643 & 0.675 & 0.704 & 0.383 & +1.10 & -0.30 \\
\midrule

\multirow{3}{*}{YOLOv12}
& Baseline & 0.647 & 0.666 & 0.706 & 0.374 & -- & -- \\
& Multi-Continuity & 0.645 & 0.648 & 0.703 & 0.380 & -0.30 & +0.60 \\
& Differencing & 0.706 & 0.608 & 0.681 & 0.372 & -2.50 & -0.20 \\
\midrule

\multirow{3}{*}{MambaYOLO}
& Baseline & 0.685 & 0.683 & 0.735 & 0.404 & -- & -- \\
& Multi-Continuity & 0.679 & 0.679 & 0.731 & 0.413 & -0.40 & +0.90 \\
& Differencing & 0.676 & 0.692 & 0.729 & 0.404 & -0.60 & +0.00 \\
\midrule

\multirow{3}{*}{DETR}
& Baseline & 0.525 & 0.457 & 0.489 & 0.243 & -- & -- \\
& Multi-Continuity & 0.608 & 0.479 & 0.526 & 0.269 & +3.70 & +2.60 \\
& Differencing & 0.642 & 0.529 & 0.571 & 0.286 & +8.20 & +4.30 \\
\bottomrule
\end{tabular}
\end{adjustbox}
\end{table}

\begin{table}[t]
\centering
\caption{Detector-wise results on NEU-DET using 50\% of the training data.}
\label{tab:app-neudet-50}

\scriptsize
\setlength{\tabcolsep}{3pt}
\renewcommand{\arraystretch}{1.08}

\begin{adjustbox}{max width=\textwidth}
\begin{tabular}{llcccccc}
\toprule
\textbf{Detector} & \textbf{Method} &
\textbf{Precision} & \textbf{Recall} &
\textbf{mAP@0.5} & \textbf{mAP@0.5:0.95} &
\textbf{$\Delta$mAP@0.5} & \textbf{$\Delta$mAP@0.5:0.95} \\
\midrule

\multirow{3}{*}{YOLOv8}
& Baseline & 0.609 & 0.623 & 0.654 & 0.338 & -- & -- \\
& Multi-Continuity & 0.643 & 0.688 & 0.710 & 0.396 & +5.60 & +5.80 \\
& Differencing & 0.666 & 0.684 & 0.709 & 0.386 & +5.50 & +4.80 \\
\midrule

\multirow{3}{*}{YOLOv10}
& Baseline & 0.618 & 0.608 & 0.651 & 0.359 & -- & -- \\
& Multi-Continuity & 0.655 & 0.637 & 0.680 & 0.373 & +2.90 & +1.40 \\
& Differencing & 0.637 & 0.656 & 0.680 & 0.362 & +2.90 & +0.30 \\
\midrule

\multirow{3}{*}{YOLO11}
& Baseline & 0.581 & 0.633 & 0.669 & 0.355 & -- & -- \\
& Multi-Continuity & 0.607 & 0.660 & 0.690 & 0.375 & +2.10 & +2.00 \\
& Differencing & 0.592 & 0.672 & 0.673 & 0.366 & +0.40 & +1.10 \\
\midrule

\multirow{3}{*}{YOLOv12}
& Baseline & 0.612 & 0.653 & 0.676 & 0.366 & -- & -- \\
& Multi-Continuity & 0.618 & 0.661 & 0.687 & 0.373 & +1.10 & +0.70 \\
& Differencing & 0.778 & 0.601 & 0.681 & 0.372 & +0.50 & +0.60 \\
\midrule

\multirow{3}{*}{MambaYOLO}
& Baseline & 0.605 & 0.656 & 0.677 & 0.369 & -- & -- \\
& Multi-Continuity & 0.650 & 0.650 & 0.692 & 0.373 & +1.50 & +0.40 \\
& Differencing & 0.658 & 0.674 & 0.701 & 0.380 & +2.40 & +1.10 \\
\midrule

\multirow{3}{*}{DETR}
& Baseline & 0.423 & 0.544 & 0.469 & 0.232 & -- & -- \\
& Multi-Continuity & 0.583 & 0.530 & 0.532 & 0.270 & +6.30 & +3.80 \\
& Differencing & 0.566 & 0.533 & 0.554 & 0.276 & +8.50 & +4.40 \\
\bottomrule
\end{tabular}
\end{adjustbox}
\end{table}

\begin{table}[t]
\centering
\caption{Detector-wise results on NEU-DET using 25\% of the training data.}
\label{tab:app-neudet-25}

\scriptsize
\setlength{\tabcolsep}{3pt}
\renewcommand{\arraystretch}{1.08}

\begin{adjustbox}{max width=\textwidth}
\begin{tabular}{llcccccc}
\toprule
\textbf{Detector} & \textbf{Method} &
\textbf{Precision} & \textbf{Recall} &
\textbf{mAP@0.5} & \textbf{mAP@0.5:0.95} &
\textbf{$\Delta$mAP@0.5} & \textbf{$\Delta$mAP@0.5:0.95} \\
\midrule

\multirow{3}{*}{YOLOv8}
& Baseline & 0.523 & 0.514 & 0.590 & 0.287 & -- & -- \\
& Multi-Continuity & 0.598 & 0.621 & 0.660 & 0.335 & +7.00 & +4.80 \\
& Differencing & 0.600 & 0.623 & 0.655 & 0.339 & +6.50 & +5.20 \\
\midrule

\multirow{3}{*}{YOLOv10}
& Baseline & 0.566 & 0.536 & 0.571 & 0.293 & -- & -- \\
& Multi-Continuity & 0.549 & 0.571 & 0.582 & 0.298 & +1.10 & +0.50 \\
& Differencing & 0.637 & 0.656 & 0.680 & 0.362 & +10.90 & +6.90 \\
\midrule

\multirow{3}{*}{YOLO11}
& Baseline & 0.518 & 0.562 & 0.579 & 0.299 & -- & -- \\
& Multi-Continuity & 0.667 & 0.573 & 0.636 & 0.317 & +5.70 & +1.80 \\
& Differencing & 0.526 & 0.604 & 0.628 & 0.322 & +4.90 & +2.30 \\
\midrule

\multirow{3}{*}{YOLOv12}
& Baseline & 0.512 & 0.598 & 0.579 & 0.296 & -- & -- \\
& Multi-Continuity & 0.680 & 0.568 & 0.626 & 0.316 & +4.70 & +2.00 \\
& Differencing & 0.506 & 0.638 & 0.605 & 0.305 & +2.60 & +0.90 \\
\midrule

\multirow{3}{*}{MambaYOLO}
& Baseline & 0.681 & 0.523 & 0.552 & 0.255 & -- & -- \\
& Multi-Continuity & 0.686 & 0.514 & 0.565 & 0.267 & +1.30 & +1.20 \\
& Differencing & 0.583 & 0.600 & 0.624 & 0.292 & +7.20 & +3.70 \\
\midrule

\multirow{3}{*}{DETR}
& Baseline & 0.441 & 0.420 & 0.393 & 0.196 & -- & -- \\
& Multi-Continuity & 0.440 & 0.427 & 0.420 & 0.200 & +2.70 & +0.40 \\
& Differencing & 0.443 & 0.458 & 0.446 & 0.216 & +5.30 & +2.00 \\
\bottomrule
\end{tabular}
\end{adjustbox}
\end{table}

\clearpage
\setcounter{table}{0}
\renewcommand{\thetable}{B\arabic{table}}
\setcounter{figure}{0}
\renewcommand{\thefigure}{B\arabic{figure}}

\section{Additional Qualitative Results}
\label{app:qualitative-results}

We provide additional qualitative analyses to further examine the effect of the proposed continuity-driven regularization on intermediate feature representations. Specifically, we compare the spatial feature responses and local continuity discrepancies between the baseline YOLOv12 detector and its continuity-regularized counterpart.

\begin{figure}[!htbp]
\centering

\begin{minipage}{0.48\linewidth}
\centering
\includegraphics[width=\linewidth]{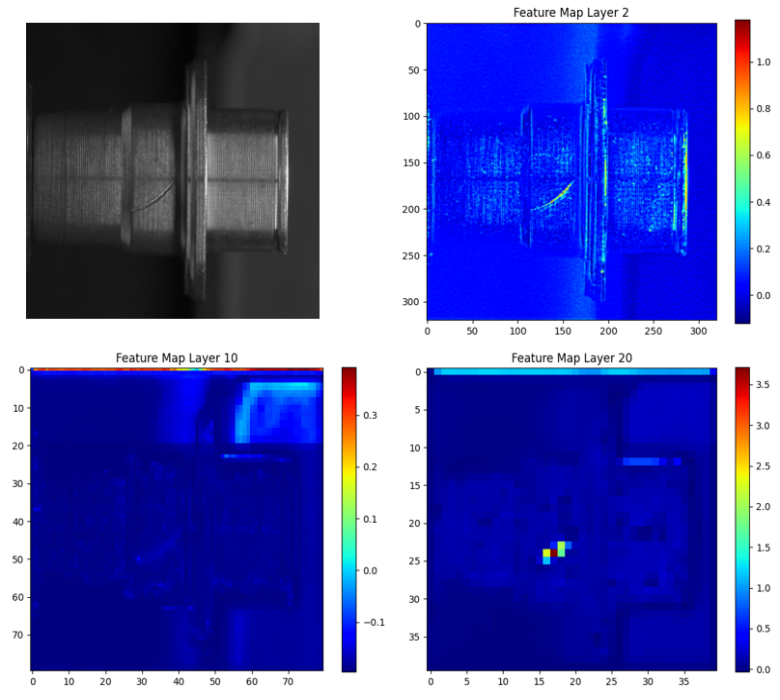}\\
(a) Baseline
\end{minipage}
\hfill
\begin{minipage}{0.48\linewidth}
\centering
\includegraphics[width=\linewidth]{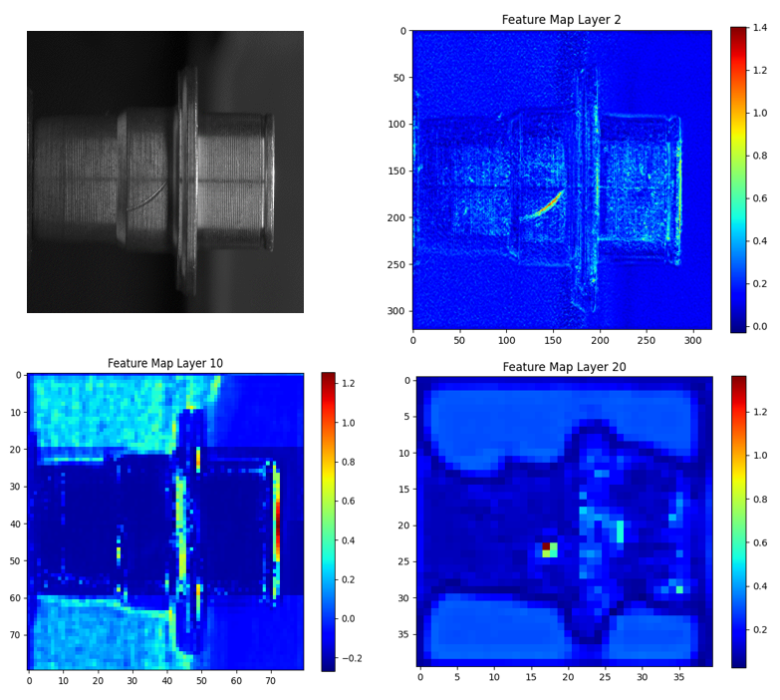}\\
(b) Continuity-regularized
\end{minipage}

\caption{Qualitative comparison of feature representations from YOLOv12.
The baseline model produces more dispersed feature responses, whereas the
proposed continuity-driven regularization yields more localized activations
around defect regions while preserving the underlying normal structures.}
\label{fig:feature-representations}
\end{figure}

\begin{figure}[!htbp]
\centering

\begin{minipage}{0.48\linewidth}
\centering
\includegraphics[width=\linewidth]{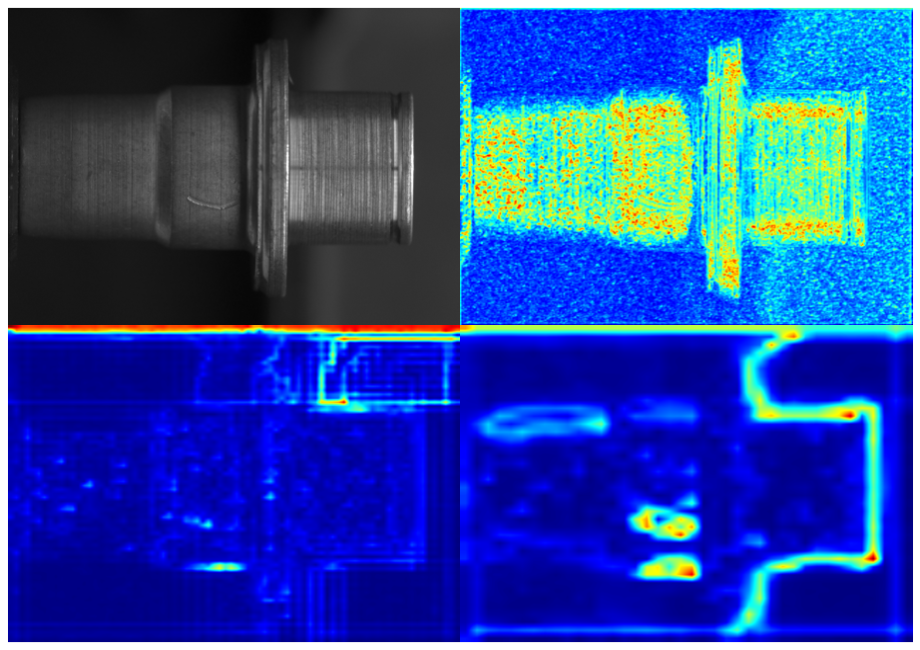}\\
(a) Baseline
\end{minipage}
\hfill
\begin{minipage}{0.48\linewidth}
\centering
\includegraphics[width=\linewidth]{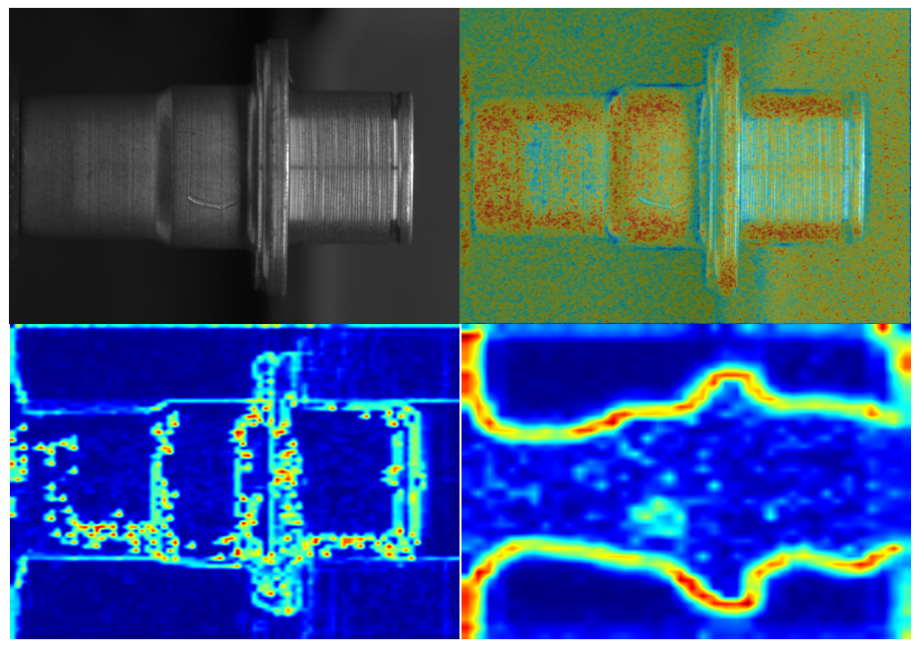}\\
(b) Continuity-regularized
\end{minipage}

\caption{Continuity discrepancy visualization from intermediate YOLOv12
feature maps. The left column shows the baseline model trained with the
standard detection loss, and the right column shows the continuity-regularized
model trained with the proposed auxiliary objective. Continuity discrepancy
maps visualize local violations of feature continuity and are used only as
diagnostic visualizations, not as detector outputs. Strong responses may occur
around defect-related disruptions as well as normal object boundaries or
abrupt structural transitions.}
\label{fig:continuity-discrepancy}
\end{figure}

\clearpage

\setcounter{table}{0}
\renewcommand{\thetable}{C\arabic{table}}
\setcounter{figure}{0}
\renewcommand{\thefigure}{C\arabic{figure}}

\section{Training Hyperparameters}
\label{app:training-hyperparameters}

This appendix summarizes the training hyperparameters used in our experiments. 
We report the method-specific regularization settings and the detector-specific training configurations for YOLO-based models and DETR.

\begin{table}[!htbp]
\centering
\caption{Method-specific hyperparameters for the proposed regularization methods.}
\label{tab:app-method-specific-hparams}
\scriptsize
\setlength{\tabcolsep}{3pt}
\renewcommand{\arraystretch}{1.08}
\begin{adjustbox}{max width=\textwidth}
\begin{tabular}{lll}
\toprule
\textbf{Method} & \textbf{Hyperparameter} & \textbf{Setting / Search Space} \\
\midrule
Baseline & Additional loss & None \\
\midrule
\multirow{3}{*}{Multi-Continuity}
& 1D continuity weight & $\beta_{\mathrm{1D}} \in \{0, 0.02, 0.04, 0.06, 0.08\}$ \\
& 2D continuity weight & $\beta_{\mathrm{2D}} \in \{0, 0.02, 0.04, 0.06, 0.08\}$ \\
& Box-region weight & $\gamma \in \{0.1, 0.2, 0.3\}$ \\
\midrule
\multirow{2}{*}{Differencing}
& First-order variation weight & $\alpha_{1} \in \{0.0, 0.1, 0.3, 0.5\}$ \\
& Second-order curvature weight & $\alpha_{2} \in \{0.0, 0.5, 1.0, 1.5, 2.0\}$ \\
\bottomrule
\end{tabular}
\end{adjustbox}
\end{table}

\begin{table}[!htbp]
\centering
\caption{Common training hyperparameters for YOLO-based detector experiments.}
\label{tab:app-yolo-common-hparams}
\scriptsize
\setlength{\tabcolsep}{3pt}
\renewcommand{\arraystretch}{1.08}
\begin{adjustbox}{max width=\textwidth}
\begin{tabular}{ll}
\toprule
\textbf{Hyperparameter} & \textbf{Setting} \\
\midrule
Model weights & YOLOv8m, YOLOv10m, YOLO11m, YOLOv12m \\
Image size & 1280 \\
Epochs & 300 \\
Early stopping patience & 20 \\
Batch size & 4 for YOLOv8/10/11; 2 for YOLOv12 \\
Learning rate schedule & $\mathrm{lr0}=0.01$, $\mathrm{lrf}=0.01$ \\
Momentum / weight decay & 0.937 / 0.0005 \\
Detection loss weights & Box = 7.5; class = 0.5; DFL = 1.5 \\
Data augmentation & Mosaic = 0.6; MixUp = 0.15; copy-paste = 0.4; scale = 0.9; horizontal flip = 0.5 \\
\bottomrule
\end{tabular}
\end{adjustbox}
\end{table}

\begin{table}[!htbp]
\centering
\caption{DETR-specific training and model hyperparameters.}
\label{tab:app-detr-hparams}
\scriptsize
\setlength{\tabcolsep}{3pt}
\renewcommand{\arraystretch}{1.08}
\begin{adjustbox}{max width=\textwidth}
\begin{tabular}{ll}
\toprule
\textbf{Hyperparameter} & \textbf{Setting} \\
\midrule
Model architecture & DETR-R50 with ResNet-50 backbone \\
Input resizing & COCO-style multi-scale resizing \\
Object queries & 200 \\
Position embedding & Sine \\
Transformer configuration & Hidden dim. = 256; encoder layers = 6; decoder layers = 6; attention heads = 8 \\
Feed-forward / dropout & FFN dim. = 2048; dropout = 0.1 \\
Auxiliary decoder loss & Enabled \\
Learning rates & Transformer/head lr = $1\times10^{-4}$; backbone lr = $1\times10^{-5}$ \\
Weight decay & $1\times10^{-4}$ \\
Epochs & 300 \\
Early stopping patience & 20 \\
Batch size & 4 \\
Hungarian matching cost & Class = 1; bbox = 5; GIoU = 2 \\
Detection loss weights & CE = 1; bbox = 5; GIoU = 2; no-object coefficient = 0.1 \\
\bottomrule
\end{tabular}
\end{adjustbox}
\end{table}

\end{document}